\documentclass[11pt,a4paper]{article}
\usepackage[hyperref]{emnlp2020}
\usepackage{times}
\usepackage{latexsym}

\usepackage{microtype}

\aclfinalcopy % Uncomment this line for the final submission
\def\aclpaperid{512} %  Enter the acl Paper ID here

\usepackage{graphicx}
\usepackage{amsmath}
\usepackage{amssymb}
\usepackage{booktabs}
\usepackage{verbatim}
\usepackage{paralist}
\usepackage{adjustbox}

\usepackage{pgfplots}
\pgfplotsset{compat=1.14}
\usepgfplotslibrary{colorbrewer}

\pgfplotsset{
    StyleLearn1/.style={mark=x, solid, Accent-H},
    StyleLearn2/.style={mark=x, solid, Accent-H},
    StyleWord1/.style={mark=triangle, solid, Accent-E},
    StyleWord2/.style={mark=triangle, solid, Accent-E},
    StyleToken1/.style={mark=o, solid, Accent-F},
    StyleToken2/.style={mark=o, solid, Accent-F},
    StyleSingle1/.style={mark=+, solid, Accent-G},
    StyleSingle2/.style={mark=+, solid, Accent-G},
    Style66a/.style={mark=x, solid, Accent-H},
    Style66b/.style={mark=o, solid, Accent-F},
    Style66c/.style={mark=otimes, solid, Set1-C},
    Style61a/.style={mark=x, solid, Accent-H},
    Style61b/.style={mark=o, solid, Accent-F},
    Style61c/.style={mark=otimes, solid, Set1-C},
    StyleA1/.style={mark=x, solid, Paired-B},
    StyleA2/.style={mark=x, solid, Paired-A},
    StyleB1/.style={mark=o, solid, Paired-F},
    StyleB2/.style={mark=o, solid, Paired-E},
    StyleC1/.style={mark=otimes, solid, Paired-D},
    StyleC2/.style={mark=otimes, solid, Paired-C},
}

\newcommand\Tstrut{\rule{0pt}{2.2ex}}
\newcommand\Bstrut{\rule[-0.1ex]{0pt}{0pt}}
\newcommand{\TBstrut}{\Tstrut\Bstrut}
\title{Fixed Encoder Self-Attention Patterns in \\ Transformer-Based Machine Translation}

\author{Alessandro Raganato, Yves Scherrer and J\"org Tiedemann\\
University of Helsinki \hspace{1em} \\
\texttt{\{name.surname\}@helsinki.fi}}

\date{}

\begin{document}
\maketitle
\begin{abstract}

Transformer-based models have brought a radical change to neural machine translation. A key feature of the Transformer architecture is the so-called multi-head attention mechanism, which allows the model to focus simultaneously on different parts of the input. However, recent works have shown that most attention heads learn simple, and often redundant, positional patterns. In this paper, we propose to replace all but one attention head of each encoder layer with simple fixed -- \textit{non-learnable} -- attentive patterns that are solely based on position and do not require any external knowledge. Our experiments with different data sizes and multiple language pairs show that fixing the attention heads on the encoder side of the Transformer at training time does not impact the translation quality and even increases BLEU scores by up to 3 points in low-resource scenarios.
\end{abstract}

\section{Introduction}

Models based on the Transformer architecture \cite{vaswani2017attention} have led to tremendous performance increases in a wide range of downstream tasks \cite{devlin-etal-2019-bert,radford2019language}, including Machine Translation (MT) \cite{vaswani2017attention,ott-etal-2018-scaling}. 
One main component of the architecture is the multi-headed attention mechanism that allows the model to %focus and
capture long-range contextual information. 
Despite these successes, the impact of the suggested parametrization choices, in particular the self-attention mechanism with its large number of attention heads distributed over several layers, has been the subject of many studies, following roughly four lines of research.

The first line of research focuses on the interpretation of the network, in particular on the analysis of attention mechanisms and the interpretability of the weights and connections %A growing body of research is dedicated to the analyses of attention mechanisms and the interpretation of weights and connections
\cite{raganato-tiedemann-2018-analysis,tang-etal-2018-self,marecek-rosa-2019-balustrades,voita-etal-2019-bottom,brunner2020identifiability}. 
A closely related research area attempts to guide the attention mechanism, e.g. by incorporating alignment objectives \cite{garg-etal-2019-jointly}, %or syntactic supervision 
or improving the representation through external information such as syntactic supervision
\cite{pham2019promoting,currey-heafield-2019-incorporating,deguchi-etal-2019-dependency}.
The third line of research argues that Transformer networks are over-parametrized and learn redundant information that can be pruned in various ways \cite{sanh2019distilbert}. For example, \newcite{voita-etal-2019-analyzing} show that a few attention heads do the ``heavy lifting'' whereas others contribute very little or nothing at all. Similarly, \newcite{Michel2019neurips} raise the question whether 16 attention heads are really necessary to obtain competitive performance. 
Finally, several recent works address the computational challenge of modeling very long sequences and modify the Transformer architecture with attention operations that reduce time complexity \cite{shen2018bidirectional,wu2019pay,child2019generating,sukhbaatar-etal-2019-adaptive,dai-etal-2019-transformer,indurthi-etal-2019-look,kitaev2020reformer}. %beltagy2020longformer 
% YS: I'm not quite sure how the fourth line is relevant for our paper: our approach doesn't care too much about long sentences nor about time complexity...

This study falls into the third category and is motivated by the observation that most self-attention patterns learned by the Transformer architecture merely reflect positional encoding of contextual information \cite{raganato-tiedemann-2018-analysis,kovaleva-etal-2019-revealing,voita-etal-2019-bottom,correia-etal-2019-adaptively}. From this standpoint, we argue that most attentive connections in the encoder do not need to be learned at all, but can be replaced by simple predefined patterns.
To this end, we design, analyze and experimentally compare intuitive and simple fixed attention patterns. The proposed patterns are solely based on positional information and do not require any learnable parameters nor external knowledge. The fixed patterns reflect %the intuition on 
the importance of locality and pose the question whether encoder self-attention needs to be learned at all % we need to learn the attentive patterns at all to be able 
to achieve state-of-the-art results in machine translation.

This paper provides the following contributions:
\begin{itemize}
\item We propose fixed -- \textit{non-learnable} -- attentive patterns that replace the learnable attention matrices in the encoder of the Transformer model. 
\item We evaluate the proposed fixed patterns on a series of experiments with different language pairs and varying amounts of training data. The results show that fixed self-attention patterns yield consistently competitive results, especially in low-resource scenarios.
\item We provide an ablation study to analyze the relative impact of the different fixed attention heads and the effect of keeping one of the eight heads learnable. Moreover, we also study the effect of the number of encoder and decoder layers on translation quality. 
\item We assess the translation performance of the fixed attention models through various contrastive test suites, focusing on two linguistic phenomena: subject-verb agreement and word sense disambiguation. %showing that having one attention head learnable focus to disambiguate. <- should we add a sentence on the findings?
% well, for that we'd have to show numbers for the 8F model... I would rather say that fixed attention works fine provided that the decoder is strong enough
\end{itemize}

Our results show that the encoder self-attention in Transformer-based machine translation can be simplified substantially, reducing the parameter footprint without loss of translation quality, and even improving quality in low-resource scenarios.

%The work presented here is most closely related to \newcite{acl2020paper}, who propose to replace learned attention by hard-coded Gaussian attention distributions. Our approach is complementary, but differs in several respects: we focus only on encoder self-attention, but on the other hand extend the number of attention heads to include additional known properties of self-attention. Furthermore, we experiment with various low-resource and high-resource scenarios and provide an ablation study to investigate the impact of the different heads.

Along with our contributions, we highlight our key findings that give insights for the further development of more lightweight neural networks while retaining state-of-the-art performance for MT: 
\begin{itemize}
\item The encoder can be substantially simplified with trivial attentive patterns at training time: only preserving adjacent and previous tokens is necessary. %TODO: should we say something about syntax heads are not needed?
% well, for that we'd have to implement them and show that we can ablate them - that sounds too complicated
\item Encoder attention heads solely based on locality principles may hamper the extraction of global semantic features beneficial for the word sense disambiguation capability of the MT model. Keeping one learnable head in the encoder compensates for degradations, but this trade-off needs to be carefully assessed.
\item Position-wise attentive patterns play a key role in low-resource scenarios, both for related (German $\leftrightarrow$ English) and unrelated (Vietnamese $\leftrightarrow$ English) languages.
\end{itemize}

\section{Related work}

Attention mechanisms in Neural Machine Translation (NMT) were first introduced in combination with Recurrent Neural Networks (RNNs) \cite{cho-etal-2014-learning,bahdanau2015neural,luong-etal-2015-effective}, between the encoder and decoder. The Transformer architecture extended the mechanism by introducing the so-called self-attention to replace the RNNs in the encoder and decoder, and by using multiple attention heads \cite{vaswani2017attention}. 
%multi-headed attention, computed over several \textit{heads} replacing the RNNs as encoder and decoder, and becoming 
This architecture rapidly became the \textit{de facto} state-of-the-art architecture for NMT, and more recently for language modeling \cite{radford2018improving} and other downstream tasks \cite{strubell-etal-2018-linguistically,devlin-etal-2019-bert,bosselut-etal-2019-comet}. 
The Transformer allows the attention for a token to be spread over the entire input sequence, multiple times, intuitively capturing different properties. This characteristic has led to a line of research focusing on the interpretation of Transformer-based networks and their attention mechanisms \cite{raganato-tiedemann-2018-analysis,tang-etal-2018-self,marecek-rosa-2019-balustrades,voita-etal-2019-bottom,vig-belinkov-2019-analyzing,clark-etal-2019-bert,kovaleva-etal-2019-revealing,tenney-etal-2019-bert,lin-etal-2019-open,jawahar-etal-2019-bert,van-schijndel-etal-2019-quantity,hao-etal-2019-visualizing,rogers2020primer}. As regards MT, recent work \cite{voita-etal-2019-analyzing} suggests that only a few attention \textit{heads} are specialized towards a specific role, e.g., focusing on a syntactic dependency relation or on rare words, and significantly contribute to the translation performance, while all the others are dispensable. 
%As mentioned above, a popular line of research focuses on the interpretation of Transformer-based networks and their attention mechanisms \cite{raganato-tiedemann-2018-analysis,tang-etal-2018-self,marecek-rosa-2019-balustrades,voita-etal-2019-bottom}.
%Another line of research argues that Transformer networks are over-parametrized and learn redundant information. For example, \newcite{voita-etal-2019-analyzing} show that a few attention heads do the ``heavy lifting'' whereas others contribute very little or nothing at all. Similarly, \newcite{Michel2019neurips} raise the question whether 16 attention heads are really necessary to obtain competitive performance.
%Two main observations have been made: First, the self-attention computation formulated in the original Transformer allows the attention for a token to be spread over the entire input sequence. This is contrary to linguistic evidence that suggests that attention is most useful on a local level, e.g. for translation of phrases. More worryingly, despite the mathematical possibility, the ability to model long-range dependencies has been questioned \cite{tang-etal-2018-self}. Second, several attention heads encode the same type of information and are redundant. \textcolor{red}{How do these observations relate to the two lines of research above?}

At the same time, recent research has attempted to bring the mathematical formulation of self-attention more in line with the linguistic expectation that attention would be most useful within a narrow local scope, e.g. for the translation of phrases \cite{hao-etal-2019-multi}. For instance, \newcite{shaw-etal-2018-self} replace the sinusoidal position encoding in the Transformer with relative position encoding to improve the focus on local positional patterns. %\newcite{yang-etal-2018-modeling} and \newcite{xu-etal-2019-leveraging}
Several studies modify the attention formula to bias the attention weights towards local areas \cite{yang-etal-2018-modeling,xu-etal-2019-leveraging,fonollosa2019joint}. \newcite{wu2019pay} and \newcite{yang-etal-2019-convolutional} use convolutional modules to replace parts of self-attention, making the overall networks computationally more efficient. \newcite{cui-etal-2019-mixed} mask out certain tokens when computing attention, which favors local attention patterns and prevents redundancy in the different attention heads.
All these contributions have shown the importance of localness, and the possibility to use lightweight convolutional networks to reduce the number of parameters while yielding competitive results \cite{wu2019pay}.
In this respect, our work is orthogonal to previous research: 
%In contrast, we do not change the architecture with other 
%we use the Transformer architecture as it is while hard coding not-learnable attentive pattern, reducing the number of parameters.
we focus only on the original Transformer architecture and investigate the replacement of learnable encoder self-attention by fixed, non-learnable attentive patterns. %use of fixed attentive patterns replacing the respective learnable components.

Recent analysis papers have identified a certain number of functions to which different self-attention heads tend to specialize. \newcite{voita-etal-2019-analyzing} identifies three types of heads: \textit{positional heads} point to an adjacent token, \textit{syntactic heads} point to tokens in a specific syntactic relation, and \textit{rare word heads} point to the least frequent tokens in a sentence. \newcite{correia-etal-2019-adaptively} identify two additional types of heads: \textit{BPE-merging heads} spread weight over adjacent tokens that are part of the same BPE cluster or hyphenated words, and \textit{interrogation heads} point to question marks at the end of the sentence. 
%We draw on these findings to design our fixed attention patterns.
In line with these findings, we design our fixed attention patterns and train NMT models without the need of learning them. %TODO: it doesnt sound super, but i was trying to add the "novel" thing that is training NMT models with our fixed patterns... 

%While \newcite{acl2020paper} only focus on positional heads, we extend their work to include BPE-merging heads and end-of-sentence heads.
In concurrent work, \newcite{acl2020paper} propose to replace learnable attention weights in Transformer-based NMT with hard-coded Gaussian distributions. This paper is complementary and differs in several respects: while \newcite{acl2020paper} consider three fixed patterns across the encoder-decoder architecture, we focus only on the encoder self-attention but present seven fixed patterns that cover additional known properties of self-attention. We study the relative impact of each of them and analyze their performance with respect to different numbers of encoder-decoder layers, and as semantic feature extractor for lexical ambiguity phenomena. Furthermore, in contrast to \newcite{acl2020paper}, we show that our fixed patterns have a clear beneficial effect in low-resource scenarios.

%\cite{ramanujan2019whats}

\begin{figure*}
\newcommand{\scalefactor}{0.092}
\centering
\includegraphics[scale=\scalefactor, clip, trim=250 0 300 0]{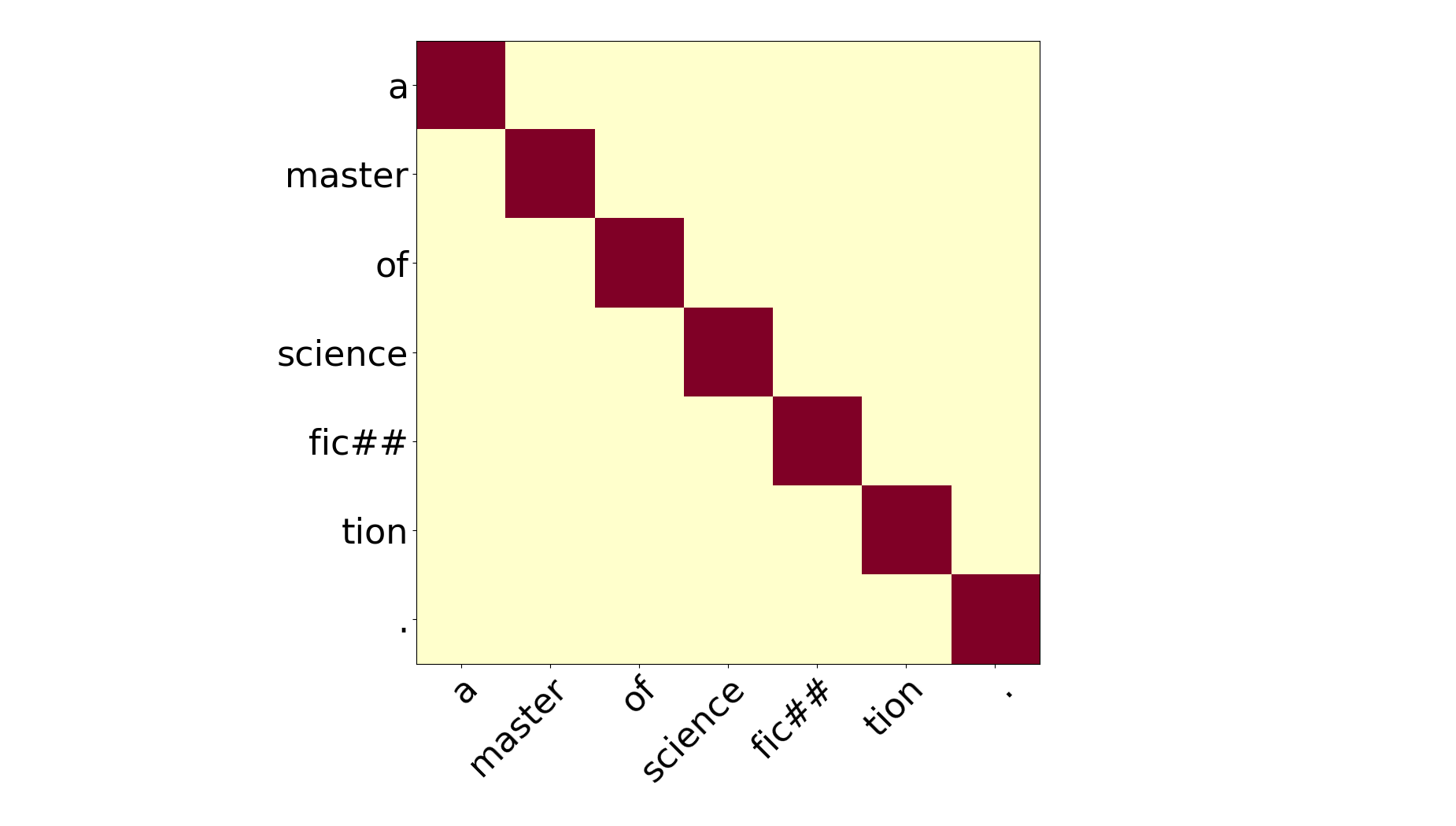}
\includegraphics[scale=\scalefactor, clip, trim=375 0 300 0]{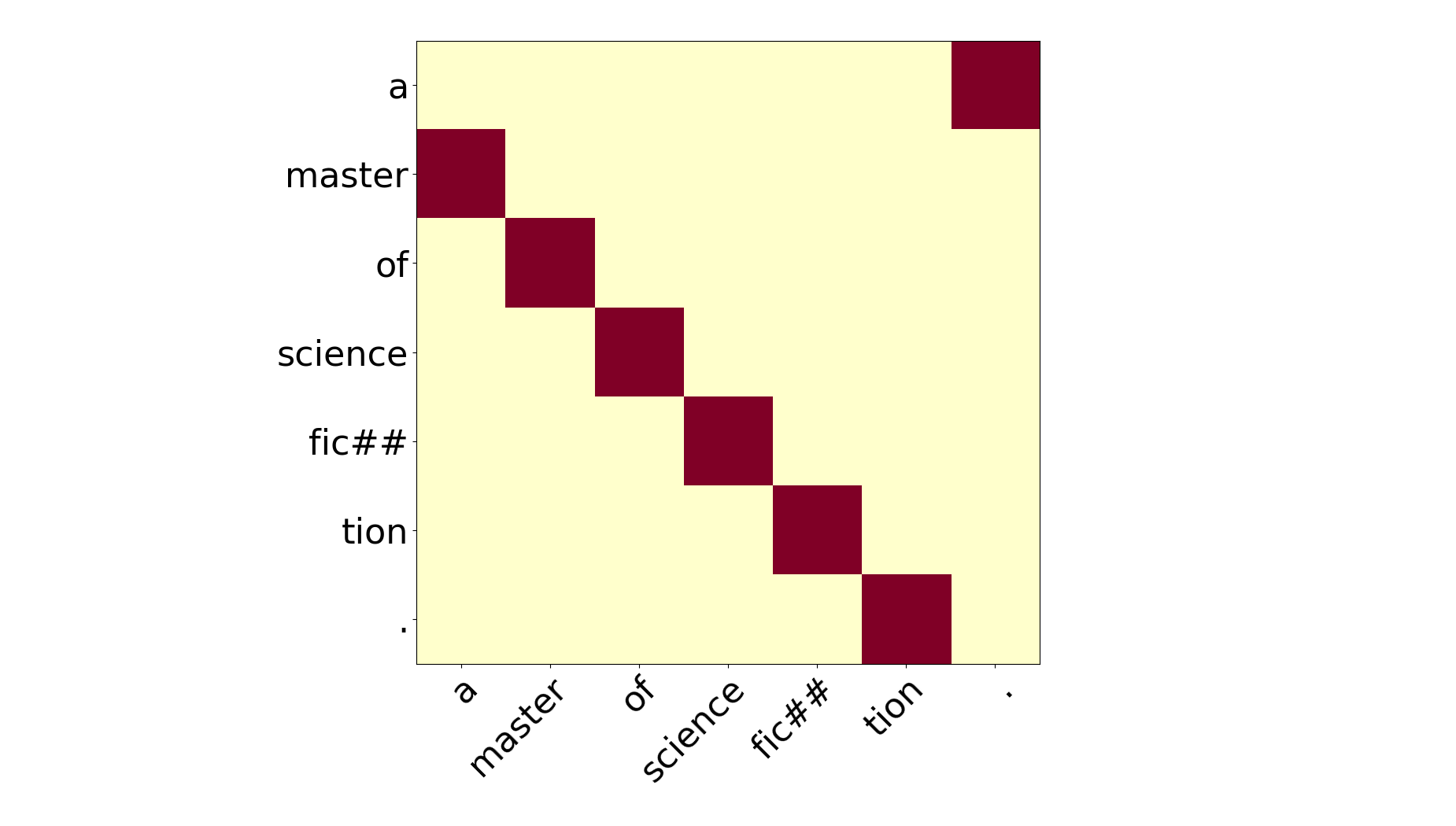}
\includegraphics[scale=\scalefactor, clip, trim=375 0 300 0]{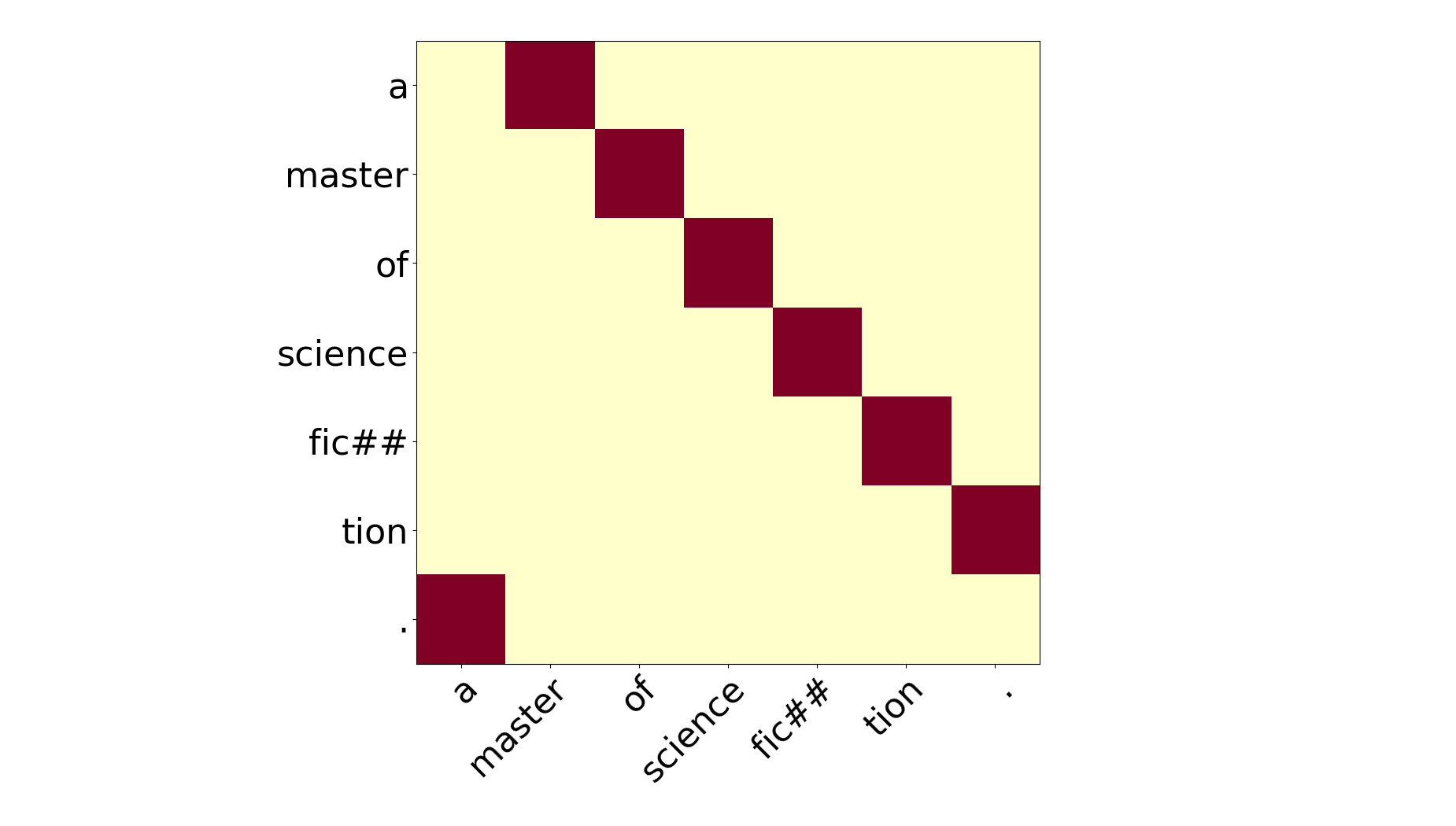}
\includegraphics[scale=\scalefactor, clip, trim=375 0 300 0]{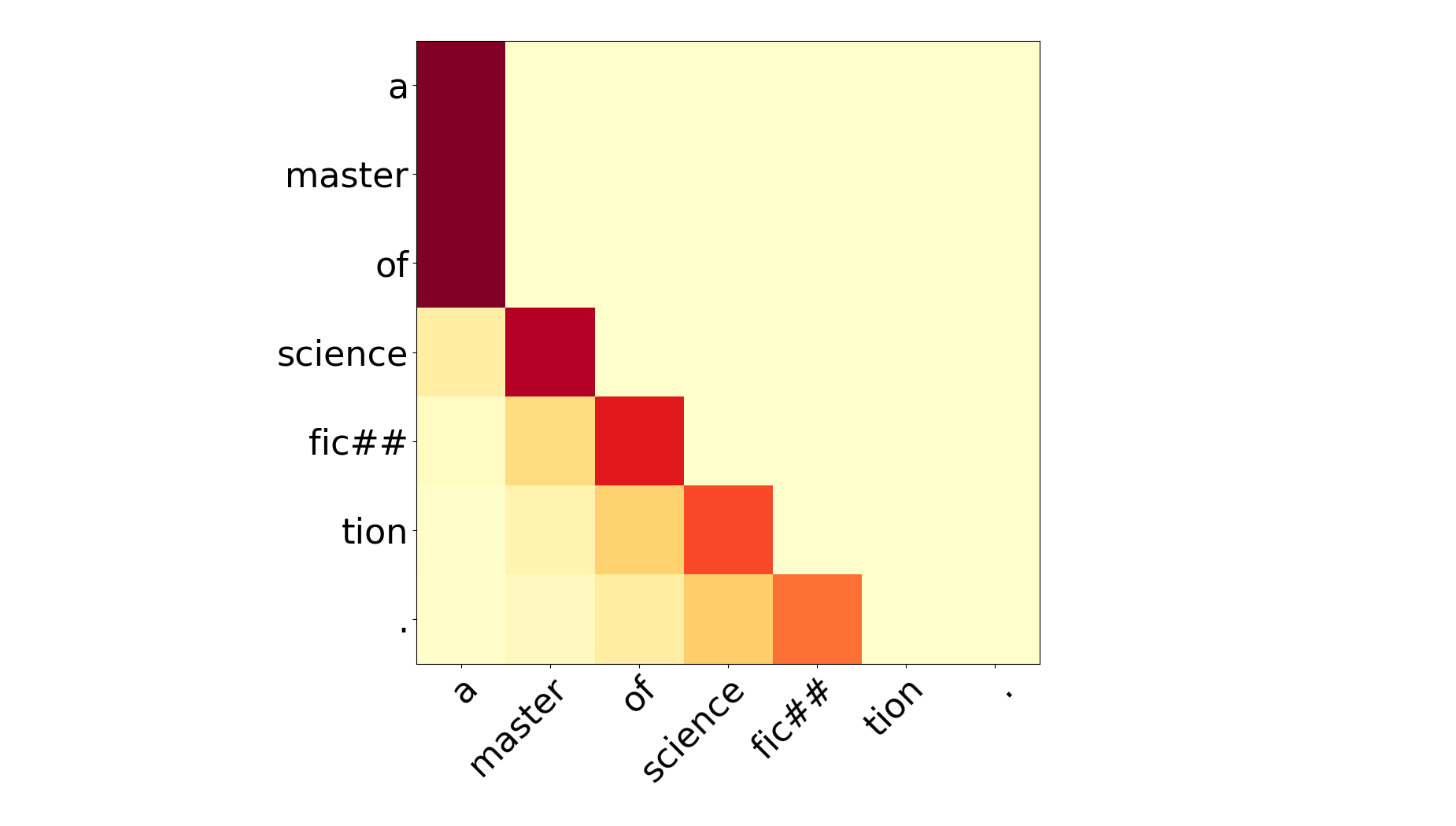}
\includegraphics[scale=\scalefactor, clip, trim=375 0 300 0]{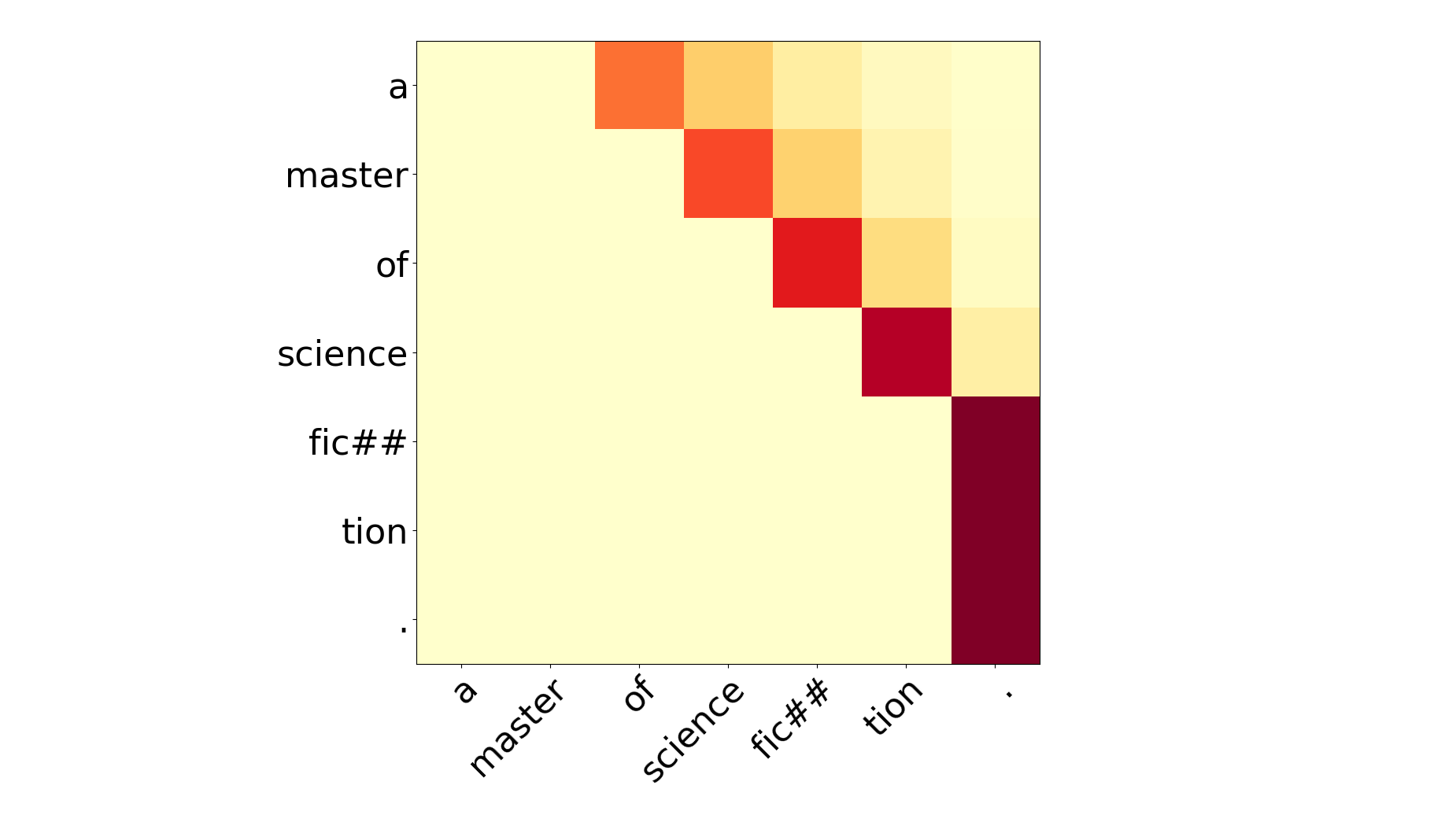}
\includegraphics[scale=\scalefactor, clip, trim=375 0 300 0]{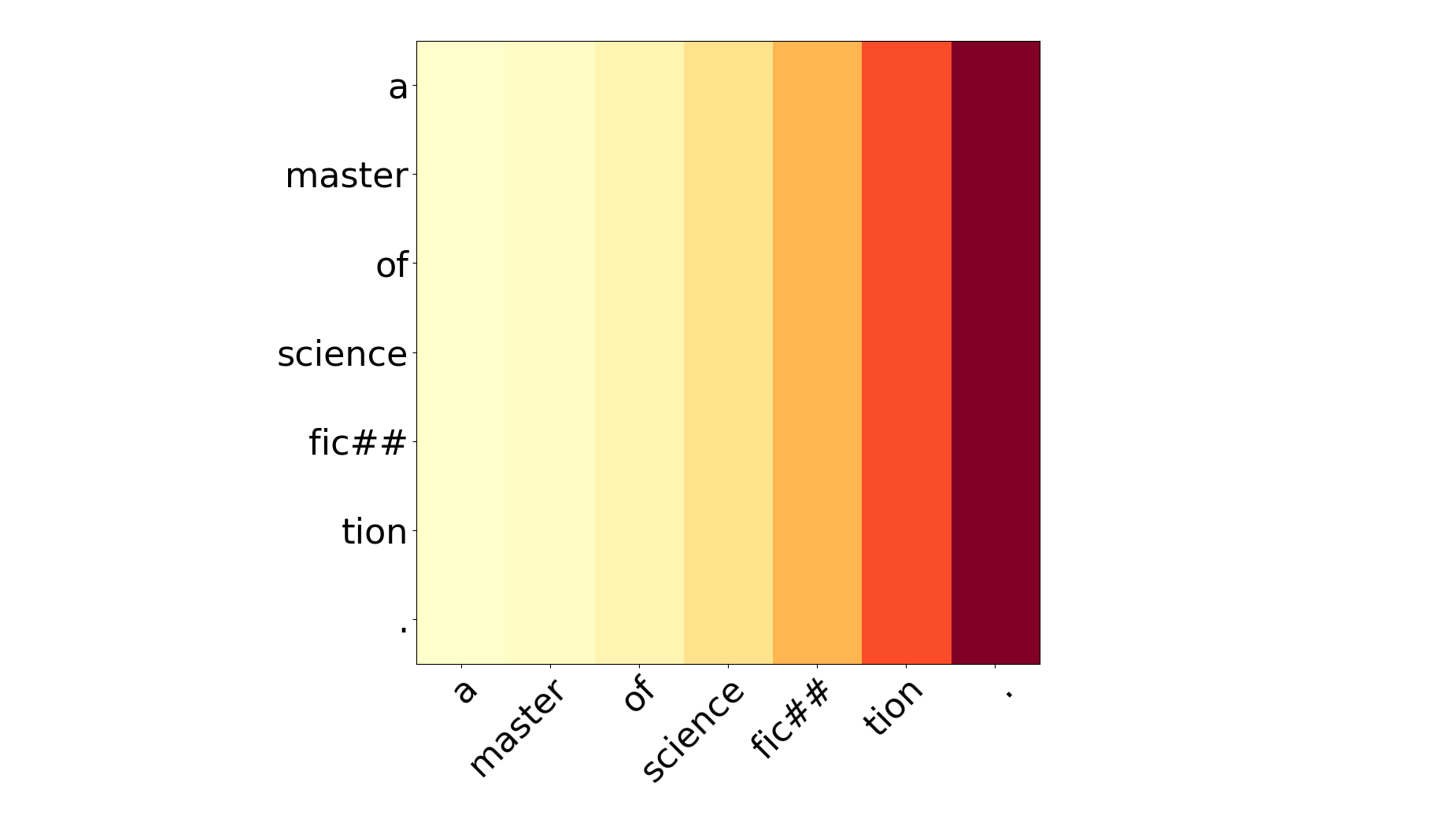}
\includegraphics[scale=\scalefactor, clip, trim=375 0 300 0]{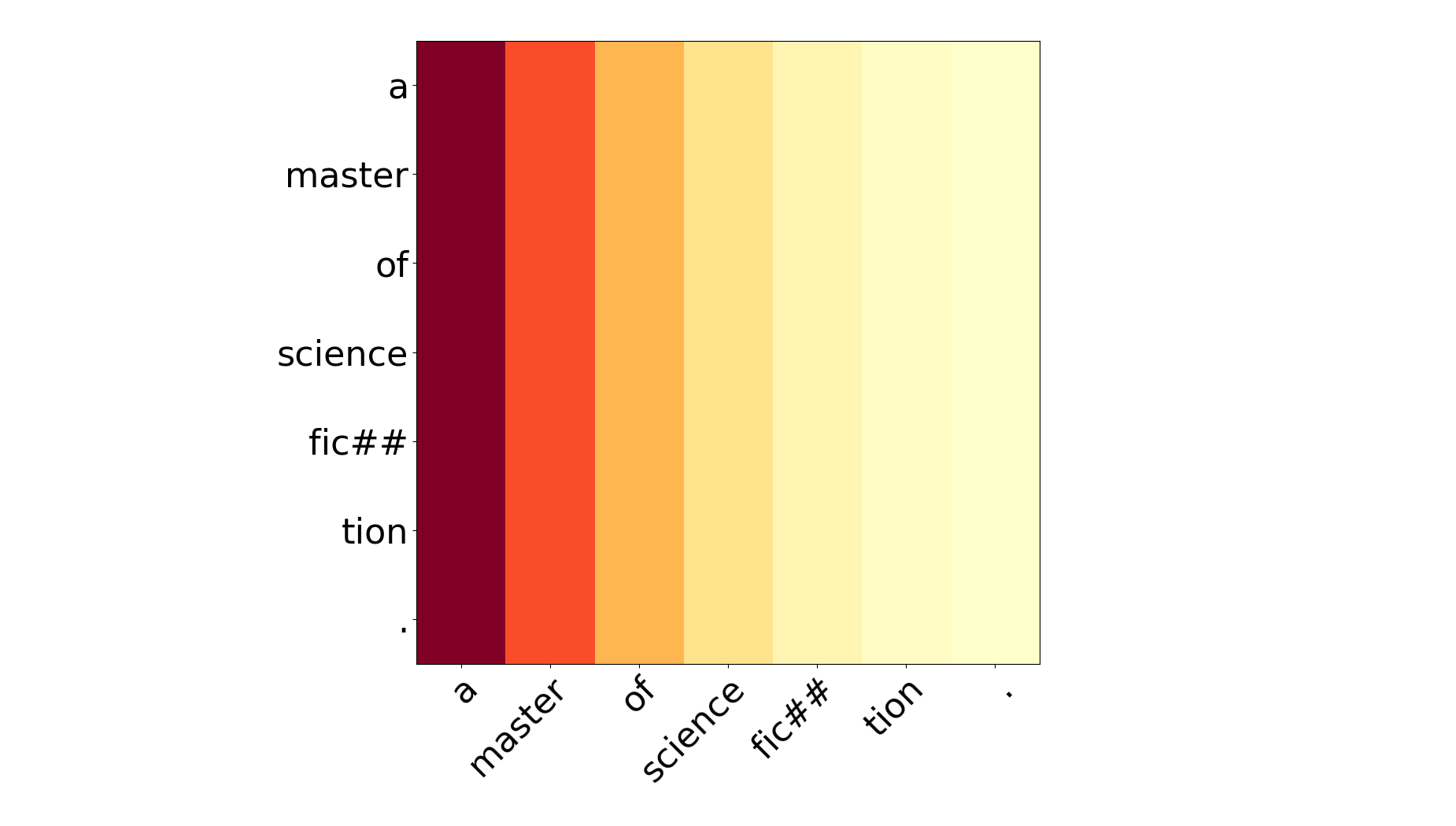}

\includegraphics[scale=\scalefactor, clip, trim=250 0 300 0]{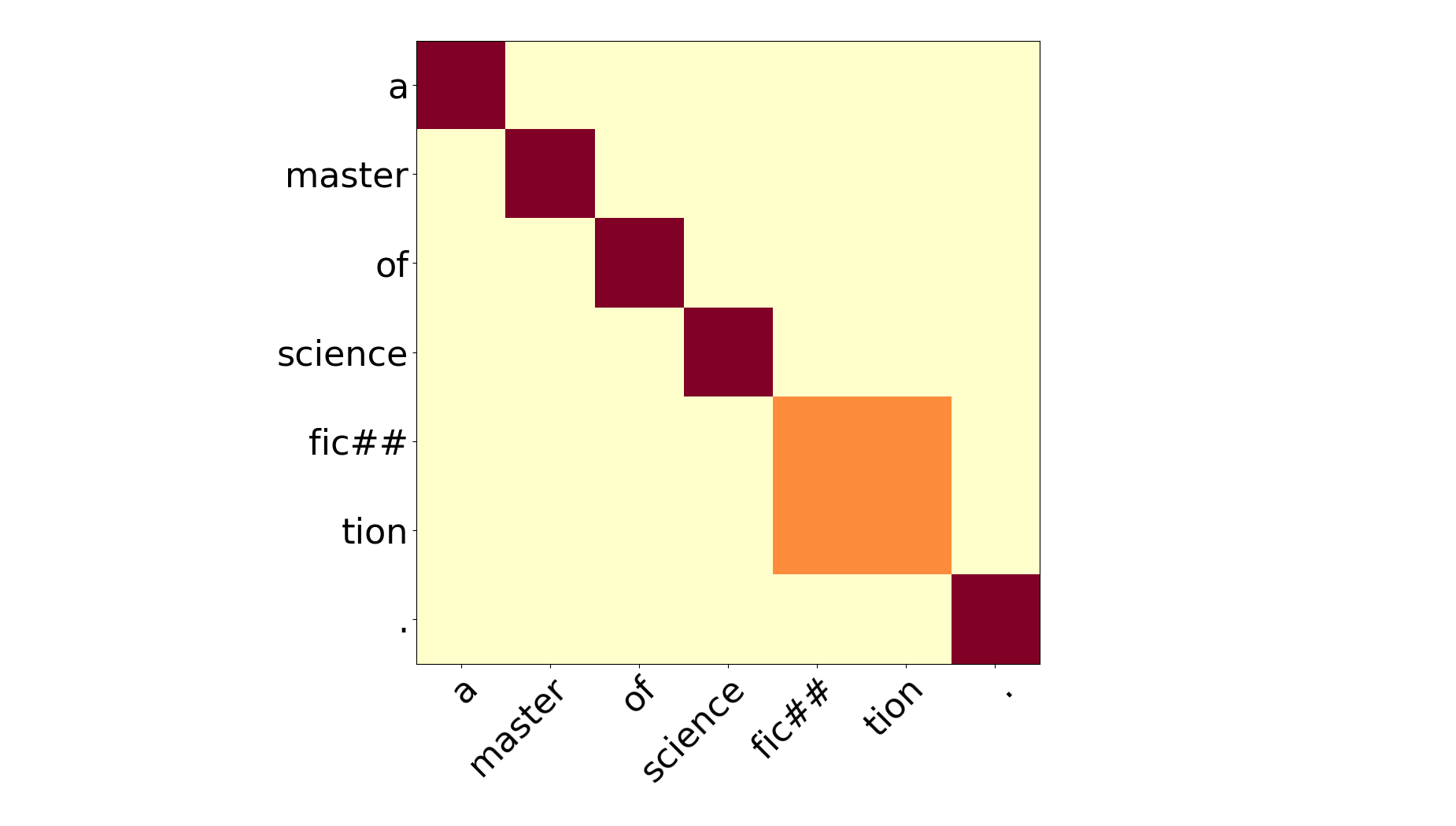}
\includegraphics[scale=\scalefactor, clip, trim=375 0 300 0]{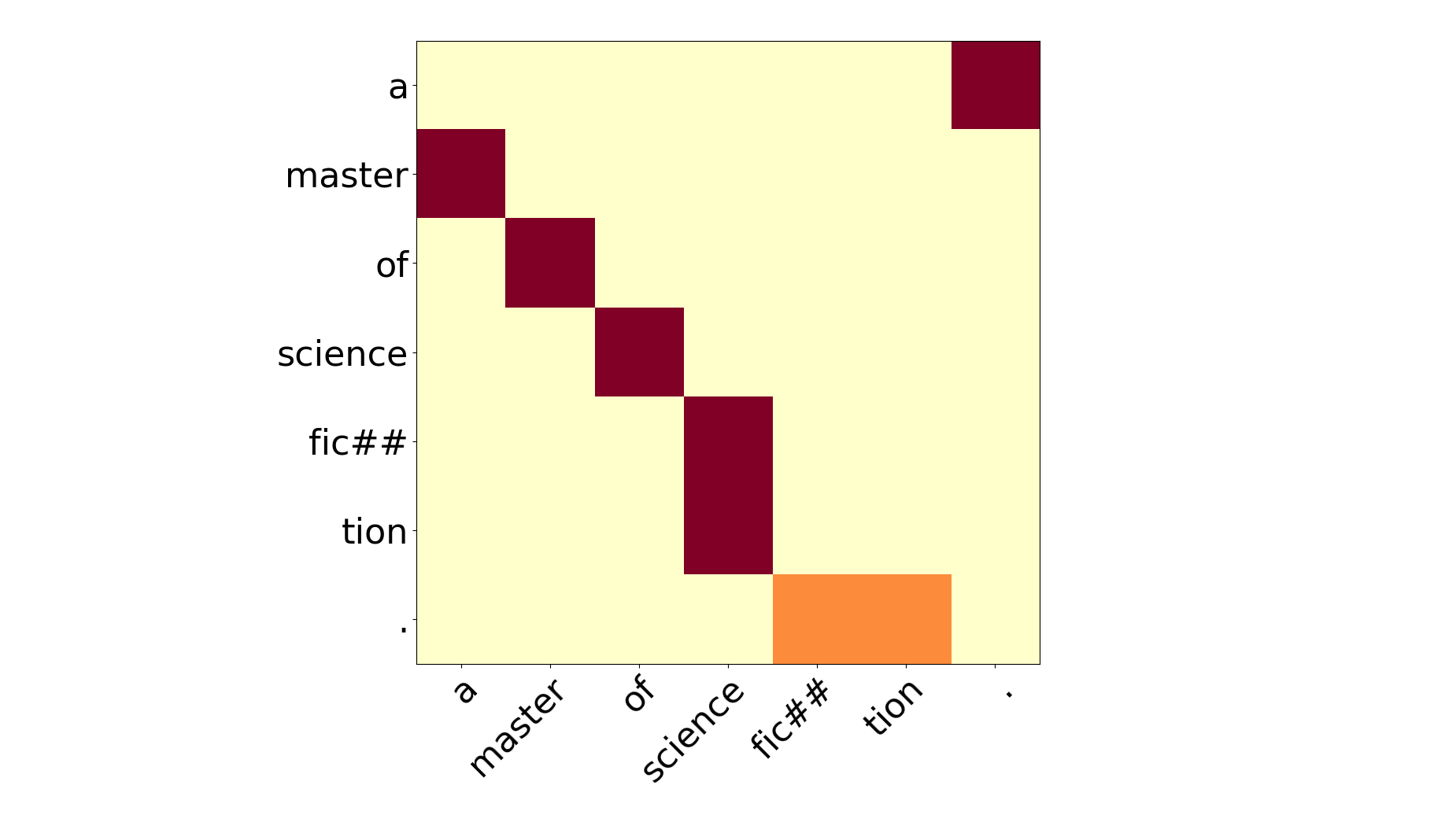}
\includegraphics[scale=\scalefactor, clip, trim=375 0 300 0]{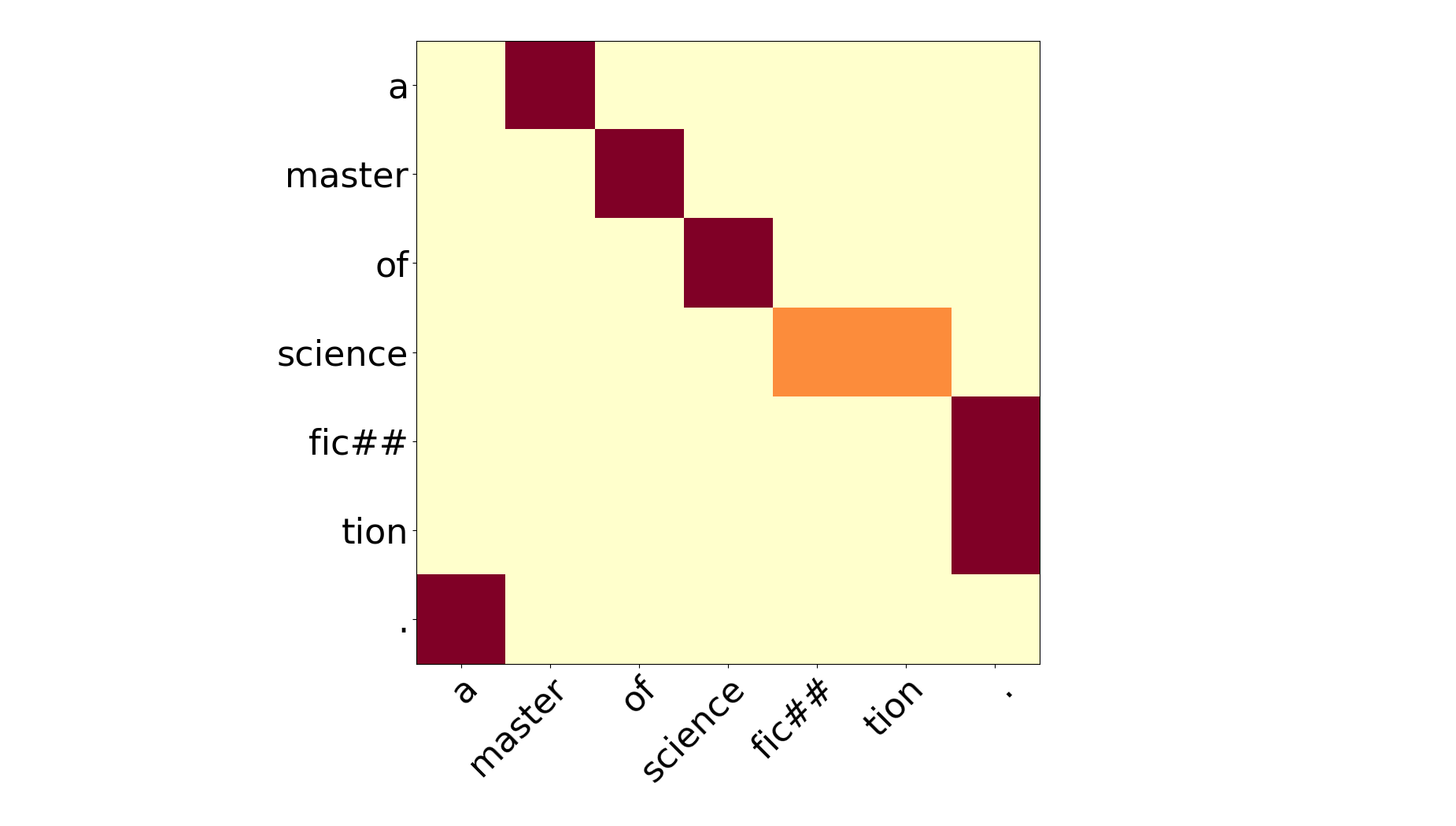}
\includegraphics[scale=\scalefactor, clip, trim=375 0 300 0]{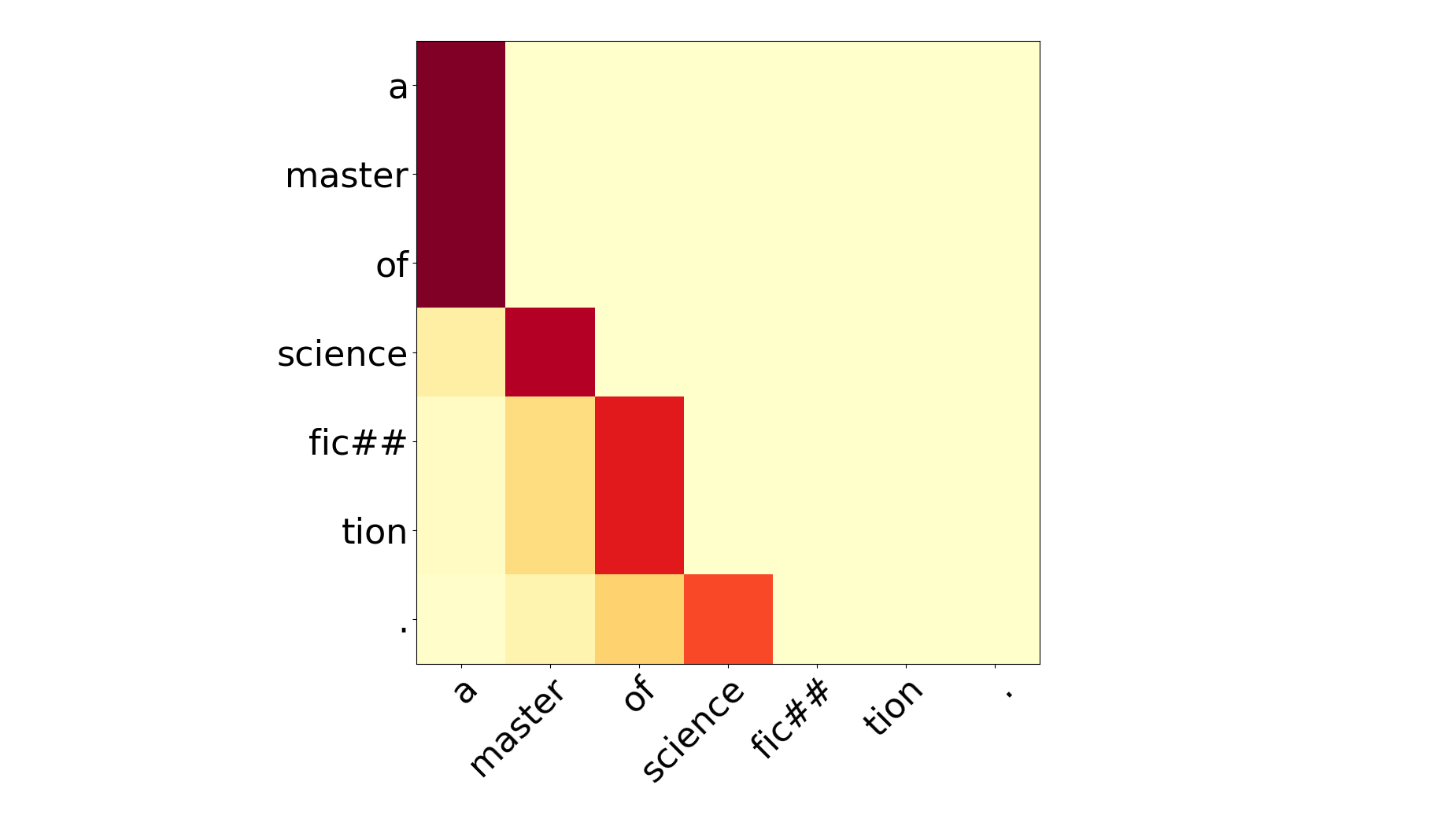}
\includegraphics[scale=\scalefactor, clip, trim=375 0 300 0]{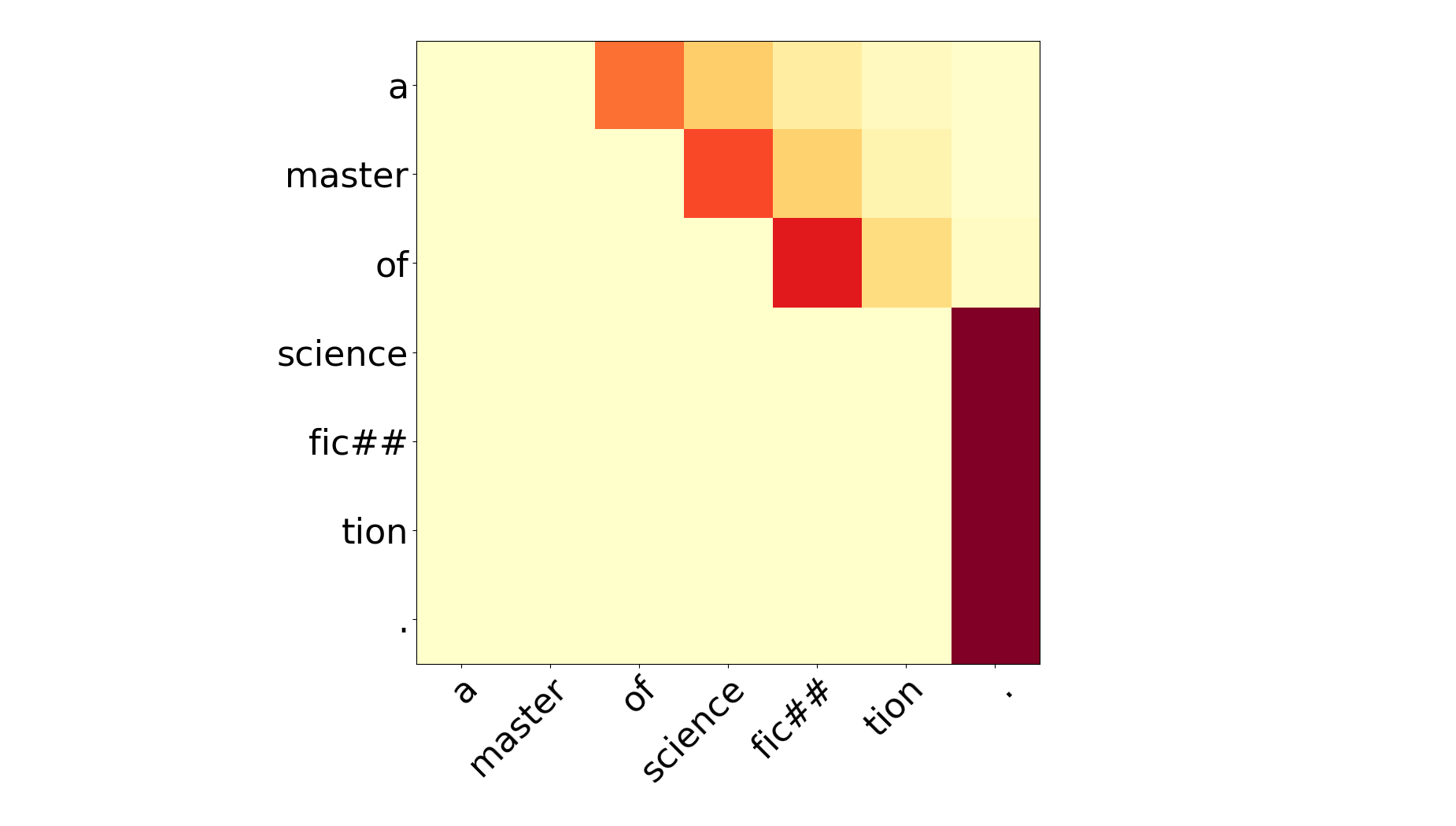}
\includegraphics[scale=\scalefactor, clip, trim=375 0 300 0]{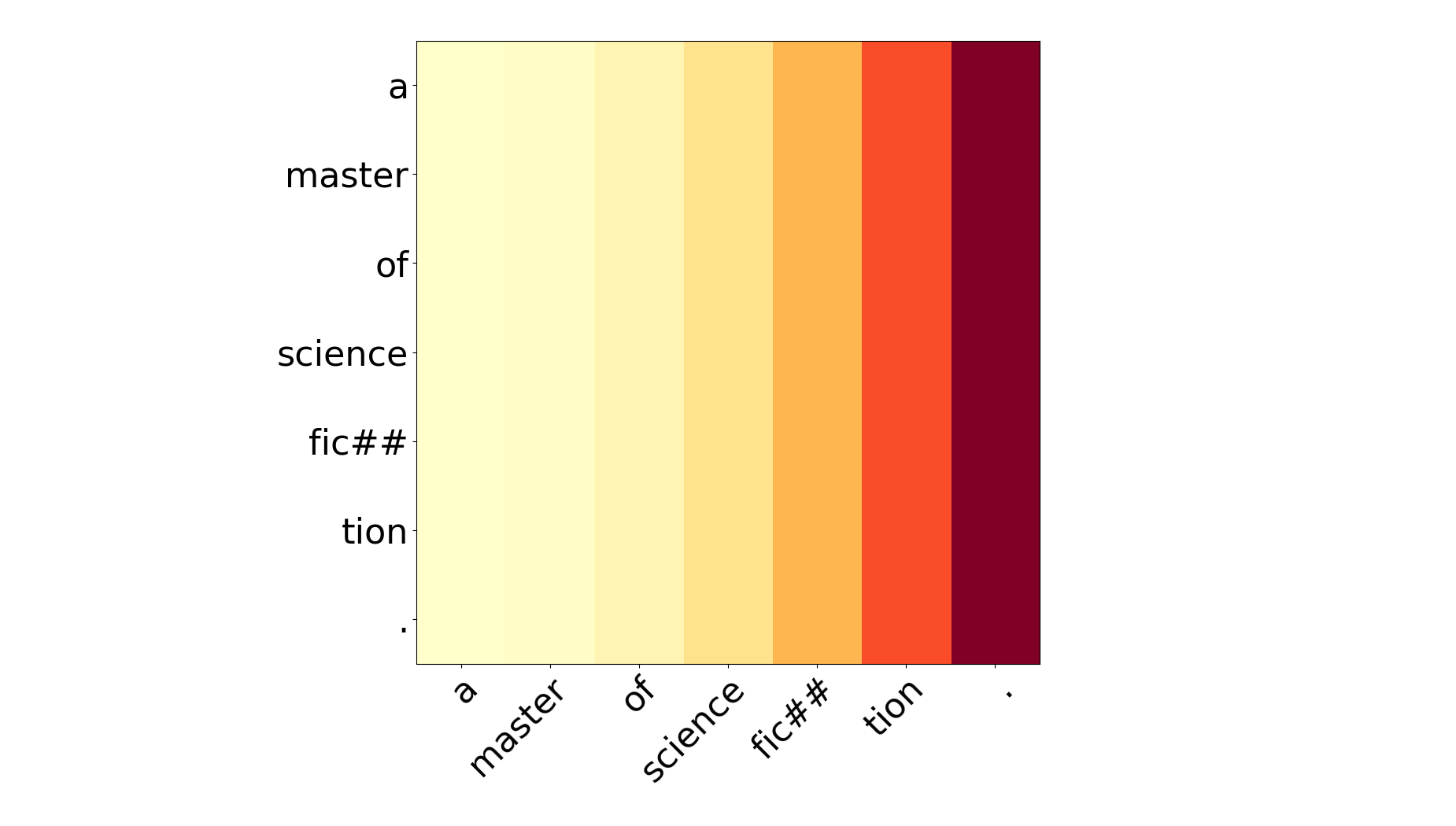}
\includegraphics[scale=\scalefactor, clip, trim=375 0 300 0]{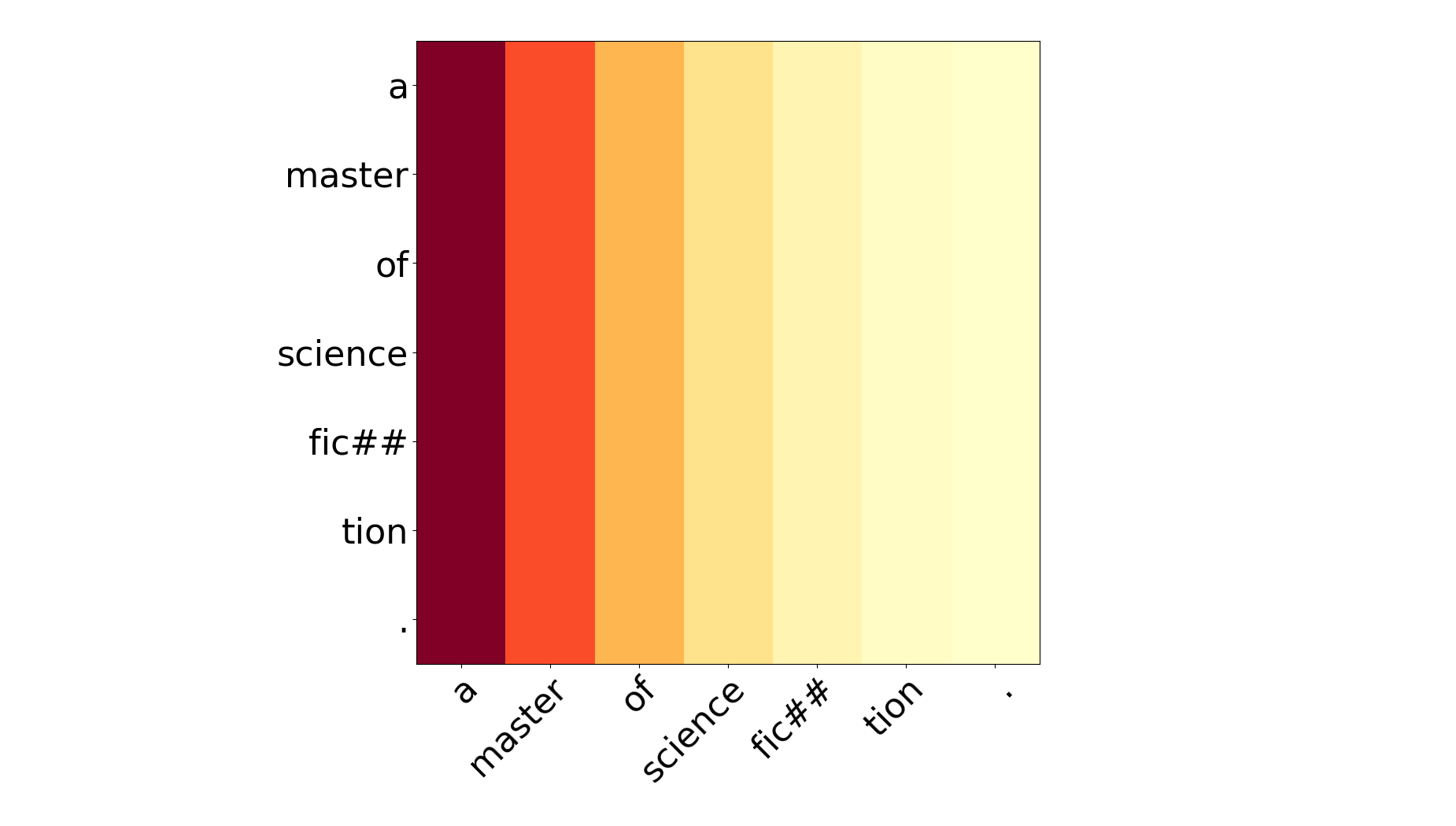}
\caption{Token-based (upper row) and word-based (lower row) fixed attention patterns for the example sentence \textit{``a master of science fic\#\# tion .''}. The word-based patterns treat the subwords ``\textit{fic\#\#}'' and ``\textit{tion}'' as a single token.}
\label{fig:attheads}
\end{figure*}

\section{Methodology}

%first
In this section, we briefly describe the Transformer architecture and its self-attention mechanism, and introduce the fixed attention patterns.% used throughout the paper.
%then

\subsection{Self-attention in Transformers}

The Transformer architecture follows the so-called encoder-decoder paradigm where the source sentence is encoded in a number of stacked encoder blocks, and the target sentence is generated through a number of stacked decoder blocks. 
Each encoder block consists of a multi-head self-attention layer and a feed-forward layer.

For a sequence of token representations $\mathbf{H} \in \mathbb{R}^{n \times d}$ (with sequence length $n$ and dimensionality $d$), the self-attention model first projects them into queries $\mathbf{Q} \in \mathbb{R}^{n \times d}$, keys $\mathbf{K} \in \mathbb{R}^{n \times d}$ and values $\mathbf{V} \in \mathbb{R}^{n \times d}$, using three different linear projections. Then, the attention energy $\xi_i$ for position $i$ in the sequence is computed by taking the scaled dot product between the query vector $\mathbf{Q}_i$ and the key matrix $\mathbf{K}$:
\begin{equation}
\xi_i = \operatorname{softmax}\left( \frac{\mathbf{Q}_i \mathbf{K}^\top}{\sqrt{d}}\right) \hspace{2em} {} \in \mathbb{R}^n
\end{equation}
The attention energy is then used to compute a weighted average of the values $\mathbf{V}$:
\begin{equation}
\operatorname{Att}(\xi_i, \mathbf{V}) = \xi_i  \mathbf{V} \hspace{2em} {} \in \mathbb{R}^d
\end{equation}
For multi-head attention with $h$ heads, the query, key and value are linearly projected $h$ times to allow the model to jointly attend to information from different representations. The attention vectors of the $h$ heads are then concatenated. Finally, the resulting multi-head attention %$\operatorname{MultiHead}(Q,K,V)$ 
is fed to a feed-forward network that consists of two linear layers with a ReLU activation in between. This multi-head attention is often called \textit{encoder self-attention}, as it builds a representation of the input sentence that is attentive to itself.
%it builds an attentive representation for the input sentence itself. 

% The multi-head attention mechanism computes the so-called \textit{scaled dot-product attention} using three weight matrices: a query $Q$, a key $K$, and a value $V$:%\footnote{$d_k$ represents the dimension of the key $K$.}
% \begin{equation}
%     \operatorname{Attention}(Q,K,V) = \operatorname{softmax}\left( \dfrac{Q K^T}{\sqrt{d_k}}\right) V
%     \label{eq:attention}

% Query, key and value are linearly projected $h$ times (where $h$ is the number of heads, i.e.~$0 \leq i < h$) to allow the model to jointly attend to information from different representations:%\footnote{$h$ is the number of heads, i.e.~$0 \leq i < h$; $W^{Q}_i\in\mathbb{R}^{d_{model}\times d_k}$, $W^{K}_i \in \mathbb{R}^{d_{model}\times d_k}$, $W^{V}_i \in \mathbb{R}^{d_{model}\times d_v}$}
% $$H_i = \operatorname{Attention}(QW^{Q}_i, KW^{K}_i, VW^{V}_i)$$
% and the results of the $h$ heads are then concatenated:%\footnote{$W^{O} \in \mathbb{R}^{hd_v\times d_{model}}$}
% $$\operatorname{MultiHead}(Q,K,V) = \operatorname{Concat}(H_1, \dots, H_h)W^O$$
% with parameter matrices $W^{Q}_i\in\mathbb{R}^{d_{model}\times d_k}$, $W^{K}_i \in \mathbb{R}^{d_{model}\times d_k}$, $W^{V}_i \in \mathbb{R}^{d_{model}\times d_v}$ and $W^{O} \in \mathbb{R}^{hd_v\times d_{model}}$.

The decoder follows the same architecture as the encoder with multi-head attention mechanisms and feed-forward networks, with two main differences: i) an additional multi-head attention mechanism, called \textit{encoder-decoder attention}, connects the last encoder layer to the decoder layers, and ii) future positions are prevented from being attended to, by masking, in order to preserve the auto-regressive property of a left-to-right decoder.

The \textit{base} version of the Transformer, the standard setting for MT, uses 6 layers for both encoder and decoder and 8 attention heads in each layer. In this work, we focus on the \textit{encoder self-attention} and replace the learned attention energy $\xi$ by fixed, predefined distributions for all but one head.
%query $Q$ and key $K$ (or more precisely, the $\operatorname{softmax}$ operator and its input of Eq.~\ref{eq:attention}) with fixed attentive patterns.

\subsection{Fixed self-attention patterns}

The inspection of encoder self-attention in standard MT models yields the somewhat surprising result that positional patterns, such as ``previous token'', ``next token'', or ``last token of the sentence'', are key features across all layers and remain even after pruning most of the attention heads \cite{voita-etal-2019-bottom,voita-etal-2019-analyzing,correia-etal-2019-adaptively}.
Instead of costly learning these trivial positional patterns using millions of sentences, we choose seven predefined patterns, each of which takes the place of an attention head (see Figure~\ref{fig:attheads}, upper row).

Given the $i{\text -}th$ word within a sentence of length $n$, we define the following patterns:
\begin{compactenum}
\item the current token: a fixed attention weight of 1.0 at position $i$,
\item the previous token: a fixed attention weight of 1.0 at position $i-1$,
\item the next token: a fixed attention weight of 1.0 at position $i+1$,
\item the larger left-hand context: a function $f$ over the positions $0$ to $i-2$,   
\item the larger right-hand context:  a function $f$ over the positions $i+2$ to $n$,   
\item the end of the sentence: a function $f$ over the positions $0$ to $n$,   
\item the start of the sentence: a function $f$ over the positions $n$ to $0$.   
\end{compactenum}

For illustration, the attention energies for patterns 2 and 4 are defined formally as follows:
$$\xi_{i,j}^{(2)} = \begin{cases}
    1 & \text{if } j = i-1 \\
    0 & \text{otherwise} \\
  \end{cases}$$
$$\xi_{i,j}^{(4)} = \begin{cases}
   f^{(4)}(j) & \text{if } j \leq i-2 \\
    0 & \text{otherwise} \\
  \end{cases}$$
where
$$f^{(4)}(j) = \frac{(j+1)^3}{\sum_{j=0}^{i-2} (j+1)^3}$$
The same function is used for all patterns, changing only the respective start and end points.\footnote{Pattern 5 and 7 are flipped versions of pattern 4 and 6, respectively.}% Note that patterns 1--3 correspond to the ``Indexing'' approach of \newcite{acl2020paper}, whereas patterns 4--7 correspond to their ``original'' approach, albeit with a different function.}
These predefined attention heads are repeated over all layers of the encoder. The eighth attention head always remains learnable.\footnote{In Section~\ref{sec:8heads}, we present a contrastive system in which the eighth head is fixed as well.}

%We define $f$ as a normalized cubic function over the positions. Specifically, for each position $i$:
%$$f(i) = \frac{(i+1)^3}{\sum_{i=start}^{end} (i+1)^3}$$
%where $start$ and $end$ are defined by the respective fixed pattern.\footnote{Pattern 5 and 7 are the flipped version of pattern 4 and 6, respectively.}

It is customary in NMT to split words into subword units, and there is evidence that self-attention treats split words differently than non-split ones \cite{correia-etal-2019-adaptively}. Therefore, we propose a second variant of the predefined patterns that assigns the same attention values to all parts of the same word %treat all parts of a word as a single token 
(see lower row of Figure~\ref{fig:attheads}).

\begin{table*}
\newcommand{\worse}[1]{\textcolor{gray}{#1}}

\centering
\begin{tabular}{lllllllll}
\toprule
& & \multicolumn{7}{c}{Encoder+Decoder layers} \TBstrut\\
\cmidrule{3-9}
& Encoder heads & 1+1 & 2+1 & 3+1 & 4+1 & 5+1 & 6+1 & 6+6 \TBstrut\\
\midrule
EN--DE & 8L & 20.61 & 21.68 & 22.63 & 23.02 & 23.18 & 23.36 & 25.02 \TBstrut\\
& 7F$_{\textrm{token}}$+1L & 20.61 & 21.58 & 22.38 & 23.15 & 23.10 & 23.07 & 24.63 \TBstrut\\
& 7F$_{\textrm{word}}$+1L & \worse{19.72*} & 21.43 & \worse{21.81*} & 22.83 & 22.74 & \worse{22.88*} & 24.85 \TBstrut\\
& 1L & \worse{18.14*} & \worse{19.88*} & \worse{21.42*} & \worse{21.71*} & \worse{22.63*} & \worse{22.29*} & \worse{23.87*} \TBstrut\\
\midrule
DE--EN & 8L & 25.66 & 27.28 & 27.88 & 28.62 & 28.71 & 29.31 & 30.99 \TBstrut\\
& 7F$_{\textrm{token}}$+1L & \worse{24.90*} & 27.01 & \worse{26.84*} & \worse{28.09*} & 28.43 & \worse{28.61*} & 30.61 \TBstrut\\
& 7F$_{\textrm{word}}$+1L & \worse{25.03*} & \worse{26.72*} & \worse{27.38*} & \worse{27.78*} & \worse{27.82*} & \worse{28.40*} & 30.69 \TBstrut\\
& 1L & \worse{23.76*} & \worse{25.75*} & \worse{26.96*} & \worse{27.34*} & \worse{27.44*} & \worse{27.56*} & \worse{30.17*} \TBstrut\\
\bottomrule
\end{tabular}
\caption{BLEU scores for the German $\leftrightarrow$ English (DE $\leftrightarrow$ EN) standard scenario, for different configurations of learnable (L) and fixed (F) attention heads. Scores marked in gray with * are significantly lower than the respective \textit{8L} model scores, at $p < 0.05$. Statistical significance is computed using the {\em compare-mt} tool \cite{neubig-etal-2019-compare} with paired bootstrap resampling with 1000 resamples \cite{koehn-2004-statistical}.}
\label{tab:midresults}
\end{table*}

% \begin{figure}
% \centering
% \begin{tikzpicture}
% \begin{axis}[
% ymin=15,
% symbolic x coords={1,2,3,4,5,6,12},
% xtick=data,
% xticklabels={{1+1},{2+1},{3+1},{4+1},{5+1},{6+1},{6+6}},
% ticklabel style={font=\small},
% legend pos={south east},
% legend style={font=\small},
% legend cell align=left
% ]
% \addplot[StyleLearn1] table {
% 1	20.61
% 2	21.68
% 3	22.63
% 4	23.02
% 5	23.18
% 6	23.36
% 12	25.02
% };
% \addlegendentry{All learned};
% \addplot[StyleWord1] table {
% 1	19.72
% 2	21.43
% 3	21.81
% 4	22.83
% 5	22.74
% 6	23.31
% 12	24.85
% };
% \addlegendentry{Fixed (word-based)};
% \addplot[StyleToken1] table {
% 1	20.61
% 2	21.58
% 3	22.38
% 4	23.15
% 5	23.1
% 6	23.07
% 12	24.63
% };
% \addlegendentry{Fixed (token-based)};
% \addplot[StyleSingle1] table {
% 1	18.14
% 2	19.88
% 3	21.42
% 4	21.71
% 5	22.63
% 6	22.29
% 12	23.87
% };
% \addlegendentry{Single-head learned};

% \addplot[StyleLearn2] table {
% 1	25.66
% 2	27.28
% 3	27.88
% 4	28.62
% 5	28.71
% 6	29.31
% 12	30.99

% };
% \addplot[StyleWord2] table {
% 1	25.03
% 2	26.72
% 3	27.38
% 4	27.78
% 5	27.82
% 6	28.4
% 12	30.69
% };
% \addplot[StyleToken2] table {
% 1	24.9
% 2	27.01
% 3	26.84
% 4	28.09
% 5	28.43
% 6	28.61
% 12	30.61
% };
% \addplot[StyleSingle2] table {
% 1	23.76
% 2	25.75
% 3	26.96
% 4	27.34
% 5	27.44
% 6	27.56
% 12	30.17
% };
% \node[font=\small] at (axis cs: 2,29) {DE--EN};
% \node[font=\small] at (axis cs: 2,18) {EN--DE};
% \end{axis}
% \end{tikzpicture}
% \caption{BLEU scores for the German $\leftrightarrow$ English standard scenario. The x-axis shows different configurations of encoder and decoder layers.}
% \label{fig:results_mid}
% \end{figure}

\section{Experiments}

We perform a series of experiments to evaluate the fixed attentive encoder patterns, starting with a standard German $\leftrightarrow$ English translation setup (Section~\ref{sec:results_mid}) and then extending the scope to low-resource and high-resource scenarios (Section~\ref{sec:results_low}). We use the OpenNMT-py \cite{klein-etal-2017-opennmt} library for training, the \textit{base} version of Transformer as hyper-parameters \cite{vaswani2017attention}, and compare against the reference using sacreBLEU \cite{papineni2002bleu,post-2018-call} 
.\footnote{Signature: BLEU+case.lc+\#.1+s.exp+tok.13a+v.1.2.11.} 
% Is lower-case BLEU used in the related work as well?
% And do we actually use sacreBLEU or just compare-mt, as noted in Table 1?

%i prefer to cite both original BLEU paper and the wrapper sacreBLEU. I guess that also the related work use lowercase bleu, but i wont care much, given that our main comparisons are our models. Compare-mt is used only for statistical significance and length analsysis. Now the problem there is that compare-mt use their own internal bleu script to compare the translations..and i need to check that..

\subsection{Results: Standard scenario} 
\label{sec:results_mid}

To assess the general viability of the proposed approach and to quantify the effects of different numbers of encoder and decoder layers, we train models on a mid-sized dataset of 2.9M training sentences from the German $\leftrightarrow$ English WMT19 news translation task \cite{barrault-etal-2019-findings}, using newstest2013 and newstest2014 as development and test data, respectively. We learn truecasers and Byte-Pair Encoding (BPE) segmentation \cite{sennrich-etal-2016-neural} on the training corpus, using 35\,000 merge operations.
%We compare against the reference using sacreBLEU \cite{papineni2002bleu,post-2018-call}.\footnote{Signature: BLEU+case.lc+\#.1+s.exp+tok.13a+v.1.2.11.} 
%Additional details about hyperparameters and the data composition and split are provided in the supplementary material.
%\textcolor{red}{Where do we put the hyperparameters? e.g. training data, subword splitting, dimensionalities, training steps/stopping criterion, toolkit?}

%Statistical significance is computed using the {\em compare-mt} tool \cite{neubig-etal-2019-compare} with paired bootstrap resampling with 1000 resamples \cite{koehn-2004-statistical} p=0.05

We train four Transformer models:
\begin{compactitem}
\item 8L: all 8 attention heads in each layer are learnable,
\item 7F$_{\textrm{token}}$+1L: 7 fixed token-based attention heads and 1 learnable head per encoder layer,
\item 7F$_{\textrm{word}}$+1L: 7 fixed word-based attention patterns and 1 learnable head per encoder layer,
\item 1L: a single learnable attention head per encoder layer.
\end{compactitem}
%\footnote{By reducing the numbers of heads, we save 2.7 M parameters.}
Each model is trained in 7 configurations: 6 encoder layers with 6 decoder layers, and 1 to 6 encoder layers coupled to 1 decoder layer. BLEU scores are shown in Table~\ref{tab:midresults}. %Figure~\ref{fig:results_mid}. 

Results for the most powerful model (6+6) show that the two fixed-attention models are almost indistinguishable from the standard model, whereas the single-head model yields consistently slightly lower results. It could be argued that the 6-layer decoder is powerful enough to compensate for deficiencies due to fixed attention on the encoder side. The 6+1 configuration, which uses a single layer decoder, shows indeed a slight performance drop for German $\to$ English, but no significant difference in the opposite direction. Overall translation quality drops significantly with three and less encoder layers, but the difference between fixed and learnable attention models is statistically insignificant in most cases. The fixed attention models always outperform the model with a single learnable head, which shows that the predefined patterns are indeed helpful. The (simpler) token-based approach seems to outperform the word-based one, but with higher numbers of decoder layers the two variants are statistically equivalent.

\subsection{Results: Low-resource and high-resource scenarios}
\label{sec:results_low}

We hypothesize that fixed attentive patterns are especially useful in low-resource scenarios since intuitive properties of self-attention are directly encoded within the model, which may be hard to learn from small training datasets. %through just a few sentences.
%We assume that fixed attentive patterns are especially useful in low-resource scenarios since the number of parameters to be learned is drastically reduced, without losing the key properties of self-attention. For instance, a fixed-attention model saves 2.7M parameters compared to a fully learnable one. 
We empirically test this assumption on four translation tasks:
%\footnote{See supplementary material for additional details.}

\begin{compactitem}
\item German $\to$ English (DE$\to$EN), using the data from the IWSLT 2014 shared task \cite{cettolo2014report}. As prior work \cite{ranzato2016sequence,sennrich-zhang-2019-revisiting}, we report BLEU score on the concatenated dev sets: tst2010, tst2011, tst2012, dev2010, dev2012 (159\,000 training sentences, 7\,282 for development, and 6\,750 for testing).
\item Korean $\to$ English (KO$\to$EN), using the dataset described in \newcite{park-hong-cha:2016:PACLIC} (90\,000 training sentences, 1\,000 for development, and 2\,000 for testing).\footnote{\url{https://github.com/jungyeul/korean-parallel-corpora}}
\item Vietnamese $\leftrightarrow$ English (VI$\leftrightarrow$EN), using the data from the IWSLT 2015 shared task \cite{cettolo2015iwslt}, using tst2012 and tst2013 for development and testing, respectively (133\,000 training sentences, 1\,553 for development and 1\,268 per testing).
\end{compactitem}

Low-resource scenarios can be sensible to the choice of hyperparameters \cite{sennrich-zhang-2019-revisiting}. Hence, we apply three of the most successful adaptations to all our configurations: reduced batch size (4k $\to$ 1k tokens), increased dropout (0.1 $\to$ 0.3), and tied embeddings. %\footnote{Note however that the optimized parameters of \newcite{sennrich-zhang-2019-revisiting} refer to RNN-based architectures and do not necessarily show the same impact in Transformer-based ones.}
Sentences are BPE-encoded with 30\,000 merge operations, shared between source and target language, but independent for Korean $\to$ English.

\begin{table}
\centering
\begin{adjustbox}{max width=\linewidth}
\setlength{\tabcolsep}{4pt}
\begin{tabular}{lrrrr}
\toprule
Enc. heads & DE--EN & KO--EN & EN--VI & VI--EN \TBstrut\\
\midrule
8L & 30.86 & 6.67 & 29.85 & 26.15 \TBstrut\\
7F$_{\textrm{token}}$+1L & \bf 32.95 & 8.43 & 31.05 & \bf 29.16 \TBstrut\\
7F$_{\textrm{word}}$+1L & 32.56 & \bf 8.70 & \bf 31.15 & 28.90 \TBstrut\\
1L & 30.22 & 6.14 & 28.67 & 25.03 \TBstrut\\
\midrule
%Previous (comp.) & $^\ddagger$ 31.97 & $^\dagger$ 5.97 & $^\uplus$ 25.61  & $^\uplus$ 22.48 \\
Prior work & $^\dagger$ 33.60 & $^\dagger$ 10.37 & $^\uplus$ 27.71 & $^\uplus$ 26.15 \TBstrut\\
\bottomrule
\end{tabular}
\end{adjustbox}
\caption{BLEU scores obtained for the low-resource scenarios with 6+6 layer configuration. Results marked with $\dagger$ are taken from \newcite{sennrich-zhang-2019-revisiting}, those marked with $\uplus$ from \newcite{kudo-2018-subword}.}
\label{tab:lowres}
\end{table}

\begin{table}
\centering
\scalebox{0.93}{
\begin{tabular}{lrr}
\toprule
Encoder heads & EN--DE & DE--EN \TBstrut\\
\midrule
8L & 26.75 & \bf 34.10 \TBstrut\\
7F$_{\textrm{token}}$+1L & 26.52 & 33.50 \TBstrut\\
7F$_{\textrm{word}}$+1L & \bf 26.92 & 33.17 \TBstrut\\
1L & 26.26 & 32.91 \TBstrut\\
\bottomrule
\end{tabular}
}
\caption{BLEU scores obtained for the high-resource scenario with 6+6 layer configurations.}
\label{tab:highres}
\end{table}

Results of the 6+6 layer configurations are shown in Table~\ref{tab:lowres}.\footnote{The 6+1 models show globally lower scores, but similar relative rankings between models.}  The models using fixed attention consistently outperform the models using learned attention, by up to 3 BLEU. No clear winner between token-based and word-based fixed attention can be distinguished though.

Our English $\leftrightarrow$ Vietnamese models outperform prior work based on an RNN architecture by a large margin, but the German $\to$ English and Korean $\to$ English models remain below the heavily optimized models of \newcite{sennrich-zhang-2019-revisiting}. 
%Both types of optimization are independent of each other and could be combined.
However, we note that our goal is not to beat the state-of-the-art in a given MT setting but rather to show the performance of simple non-learnable attentive patterns across different language pairs and data sizes. %TODO: maybe to improve, but i think we should say something along this line
Moreover, it is worth to mention that the Korean $\to$ English dataset, being automatically created, includes some noise in the test data that may impact the comparison.\footnote{\url{https://github.com/jungyeul/korean-parallel-corpora/issues/1}} %However, we still include it for comparison reasons.

%\textcolor{red}{YS: How should we best address the fact that our models perform worse than Sennrich \& Zhang? Their model is an RNN, so we can't implement all of their changes, as one reviewer suggested. The "agressive dropout" seems to be very successful (p = 0.3 for dropping words; p = 0.5 for other dropout) -- is this something that can easily be added?}

Finally, we also evaluate a high-resource scenario for German $\leftrightarrow$ English with 11.5M training sentences.\footnote{This scenario uses the same benchmark from Section~\ref{sec:results_mid}, increasing the training data with a filtered version of ParaCrawl (bicleaner filtered v3).} %TODO: can we cite the opus filter paper here? and do paracrawl has a paper to be cited?
Table~\ref{tab:highres} shows that the results of the fixed attention models do not degrade even when abundant training data allow all attention heads to be learned accurately.

\subsection{Ablation study}\label{sec:abl_study}

We perform an ablation study to assess the contribution of each attention head separately. To this end, we mask out one attention pattern across all encoder layers at test time. Table~\ref{tab:ablation} shows the differences compared to the full model, on the mid-sized German $\leftrightarrow$ English and on the Vietnamese $\leftrightarrow$ English models, both in the 6+1 and 6+6 layer configurations.

\begin{table}
\centering
\begin{adjustbox}{max width=\linewidth}
\setlength{\tabcolsep}{4pt}
\begin{tabular}{lrrrr}
\toprule
Disabled head & \multicolumn{2}{c}{6+1 layers} & \multicolumn{2}{c}{6+6 layers} \TBstrut\\
\midrule
 & EN--DE & DE--EN & EN--DE & DE--EN \TBstrut\\
\cmidrule(lr){2-3} \cmidrule(lr){4-5}  
1 Current word & -0.15 & 0.11 & 0.12 & -0.04 \TBstrut\\
2 Previous word & \bf -5.72 & \bf -5.21 & \bf -3.05 & \bf -3.26 \TBstrut\\
3 Next word & \bf -1.80 & \bf -1.98 & \bf -2.08 & \bf -1.36 \TBstrut\\
4 Prev. context & \bf -4.73 & \bf -5.20 & \bf -1.42 & \bf -2.85 \TBstrut\\
5 Next context & -0.72 & -0.34 & -0.47 & -0.66 \TBstrut\\
6 Start context & -0.17 & -0.12 & 0.14 & 0.13 \TBstrut\\
7 End context & -0.02 & 0.12 & -0.30 & 0.10 \TBstrut\\
8 Learned head & \bf -2.22 & \bf -4.05 & -0.58 & -0.78 \TBstrut\\
\midrule
 & EN--VI & VI--EN & EN--VI & VI--EN \TBstrut\\
\cmidrule(lr){2-3} \cmidrule(lr){4-5} 
1 Current word  & 0.12 & -0.14 & 0.16 & -0.05 \TBstrut\\
2 Previous word & \bf -2.32 & \bf -2.67 & \bf -2.71 & \bf -3.04 \TBstrut\\
3 Next word & \bf -1.12 & \bf -1.61 & \bf -1.35 & \bf -2.15 \TBstrut\\
4 Prev. context & \bf -4.11 & \bf -4.32 & \bf -2.82 & \bf -3.09 \TBstrut\\
5 Next context & -0.27 & -0.50 & -0.83 & -0.77 \TBstrut\\
6 Start context & -0.29 & -0.08 & -0.04 & 0 \TBstrut\\
7 End context & 0.28 & -0.29 & -0.23 & -0.19 \TBstrut\\
8 Learned head & -0.57 & -0.88 & -0.18 & 0.36 \TBstrut\\
\bottomrule
\end{tabular}
\end{adjustbox}
\caption{BLEU score differences with respect to the full model with 8 enabled heads (values $<$ -1 in bold).}
\label{tab:ablation}
\end{table}

We find that heads 2, 3 and 4 (previous word, next word, previous context) are particularly important, whereas the impact of the remaining context heads is small. Head 1 (current word) is not useful in the token-based model, but shows slightly larger numbers in the word-based setting.
%TODO: should we add the figures for word-based head 1?

The most interesting results concern the eighth, learned head. Its impact is significant, but in most cases lower than the three main heads listed above. Interestingly, disabling it causes much lower degradation in the 6+6 configurations, which suggests that a more powerful decoder can compensate for the absence of learned encoder representations.

\subsection{Eight fixed heads} \label{sec:8heads}

The ablation study suggests that it is not crucial to keep one learnable head in the encoder layers, especially if the decoder is deep enough. Here, we assess the extreme scenario where the eighth attention head is fixed as well. The eighth fixed attentive pattern focuses on the last token, with a fixed weight of 1.0 at position $n$. Table~\ref{tab:8fix} shows the results for the standard English $\leftrightarrow$ German scenario and the low-resource English $\leftrightarrow$ Vietnamese scenario. 
%\textcolor{red}{How easy is it to compute significance tests for this table?}

\begin{table}
\centering
\begin{adjustbox}{max width=\linewidth}
\setlength{\tabcolsep}{4pt}
\begin{tabular}{lrrrrr}
\toprule
Enc. heads & \#Param. & EN--DE & DE--EN & EN--VI & VI--EN  \TBstrut\\
\midrule
8L & 91.7M & 25.02 & 30.99 & 29.85 & 26.15 \TBstrut\\
7F$_{\textrm{token}}$+1L & 88.9M & 24.63 & 30.61 & 31.05 & 29.16 \TBstrut\\
8F$_{\textrm{token}}$ & 88.5M & 24.64 & 30.56 & 31.45 & 28.97  \TBstrut\\
\bottomrule
\end{tabular}
\end{adjustbox}
\caption{BLEU scores for the experiments with eight fixed attention heads and 6+6 layers. ``\#Param.'' denotes the number of parameters for the EN--DE model.}% The 7F$_{\textrm{token}}$+1L results are repeated from Tables~\ref{tab:midresults} and \ref{tab:lowres}.} %TODO: comment something on the number of paramentrs on the text, maybe should we includehere also the 1AH model? just to see better the comparison..
\label{tab:8fix}
\end{table}

Overall, the learnable attention head is completely dispensable across both language pairs. As shown in Section~\ref{sec:abl_study}, the impact of having learnable attention heads on the encoder side is negligible.
Moreover, we also note that as we replace attention heads with non-learnable ones, our configurations reduce the number of parameters without degrading translation quality.%while, at the same time, achieving comparable performances of a fully learnable model. 

\section{Analysis}

To further analyze the fixed attentive encoder patterns, we perform three targeted evaluations: i) on the sentence length, ii) on the subject-verb agreement task, and iii) on the Word Sense Disambiguation (WSD) task. 
The length analysis inspects the translation quality by sentence length. 
%groups sentences of similar length together computing BLEU score for each group.  
%As length analysis, we use the \textit{compare-mt} toolkit \cite{neubig-etal-2019-compare}, computing BLEU score by reference sentence length. 
The subject-verb agreement task is commonly used to evaluate long-range dependencies \cite{linzen-etal-2016-assessing,tran-etal-2018-importance,tang-etal-2018-self}, while the WSD task addresses lexical ambiguity phenomena, i.e., words of the source language that have multiple translations in the target language representing different meanings
\cite{marvin-koehn-2018-exploring,liu-etal-2018-handling,pu-etal-2018-integrating,tang-etal-2019-encoders}.
%, i.e., translating an ambiguous word with its correct sense 
%, i.e., words of the source language that have multiple translations in the target language representing different meanings

For both tasks, we use contrastive test suites \cite{sennrich-2017-grammatical,popovic-castilho-2019-challenge} %, measuring specific linguistic phenomena relying 
that rely on the ability of NMT systems to score given translations.
Broadly speaking, a sentence containing the linguistic phenomenon of interest is paired with the correct reference translation and with a modified translation with a specific type of error. A contrast is considered successfully detected if the reference translation obtains a higher score than the artificially modified translation.
The evaluation metric corresponds to the accuracy over all decisions.

We conduct the analyses using the DE--EN models from Section~\ref{sec:results_mid}, i.e., 8L, 7F$_{\textrm{token}}$+1L, and 7F$_{\textrm{token}}$ (H8 disabled).

\subsection{Sentence length analysis}

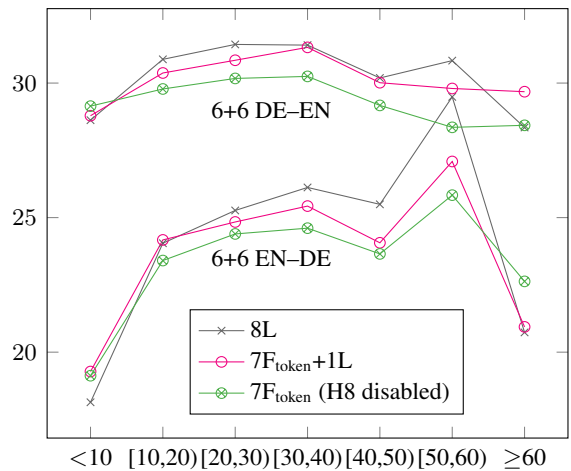
\begin{figure}
\centering
\begin{tikzpicture}
\begin{axis}[
xtick=data,
xticklabels={{$<$10},{[10,20)},{[20,30)},{[30,40)},{[40,50)},{[50,60)},{$\geq$60}},
ticklabel style={font=\small},
legend style={at={(0.8,0.3)}, font=\small},
legend cell align=left
]
\addplot[Style66a] table {
1 18.1376
2 24.0483
3 25.2657
4 26.1209
5 25.4933
6 29.4872
7 20.7341
};
\addlegendentry{8L};

\addplot[Style66b] table {
1 19.2798
2 24.1708
3 24.8381
4 25.4285
5 24.064
6 27.0863
7 20.9385
};
\addlegendentry{7F$_{\textrm{token}}$+1L};

\addplot[Style66c] table {
1 19.1249
2 23.4046
3 24.3931
4 24.6076
5 23.6513
6 25.8349
7 22.6302
};
\addlegendentry{7F$_{\textrm{token}}$ (H8 disabled)};

\addplot[Style66a] table {
1 28.6207
2 30.8818
3 31.4363
4 31.4105
5 30.1894
6 30.8344
7 28.3493
};

\addplot[Style66b] table {
1 28.7865
2 30.3747
3 30.8476
4 31.3268
5 30.0133
6 29.7986
7 29.6794
};

\addplot[Style66c] table {
1 29.1429
2 29.7805
3 30.1724
4 30.2479
5 29.1696
6 28.3549
7 28.4299
};

\node[font=\small] at (axis cs: 3.5,23.5) {6+6 EN--DE};
\node[font=\small] at (axis cs: 3.5,29) {6+6 DE--EN};
\end{axis}
\end{tikzpicture}
\caption{BLEU scores for different ranges of sentence lengths.} %\textcolor{red}{YS: How are the frequencies distributed? Would it make sense to merge the 50--60 and the 60+ categories?}
\label{fig:length}
\end{figure}

To assess whether our fixed attentive patterns may hamper modeling of global dependencies supporting long sentences, we compute BLEU score by reference sentence length.\footnote{We use the \textit{compare-mt} toolkit by \newcite{neubig-etal-2019-compare}.} Despite the small performance gap between models, as we can see from Figure~\ref{fig:length}, long sentences benefit from having \textit{learnable} attentive patterns. This is clearly shown by the 7F$_{\textrm{token}}$ (H8 disabled) model, which is consistently degraded in almost every length bin. 
%TODO look at the frequencies of the sentences in each length bin

\subsection{Subject-verb agreement}

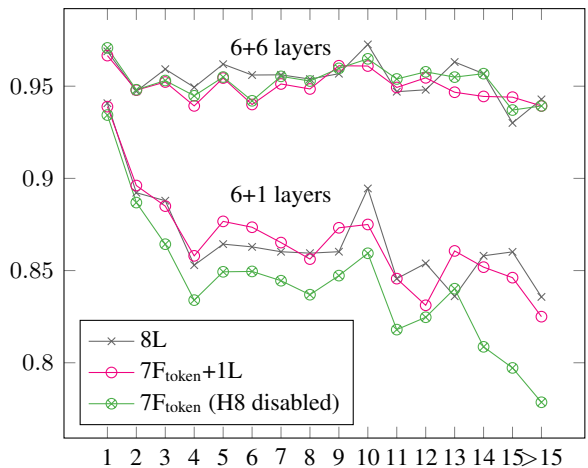
\begin{figure}
\centering
\begin{tikzpicture}
\begin{axis}[
xtick=data,
xticklabels={1,2,3,4,5,6,7,8,9,10,11,12,13,14,15,{$>$15}},
ticklabel style={font=\small},
legend pos={south west},
legend style={font=\small},
legend cell align=left
]
\addplot[Style66a] table {
1	0.968908589
2	0.94723876
3	0.959238402
4	0.9494311
5	0.961930295
6	0.956058589
7	0.956162117
8	0.953862661
9	0.956772334
10	0.97265625
11	0.947004608
12	0.948051948
13	0.963114754
14	0.956790123
15	0.93006993
16	0.942857143
};
\addlegendentry{8L};
\addplot[Style66b] table {
1	0.966596004
2	0.94785465
3	0.952266023
4	0.93931732
5	0.954423592
6	0.940079893
7	0.951199338
8	0.948497854
9	0.961095101
10	0.9609375
11	0.949308756
12	0.954545455
13	0.946721311
14	0.944444444
15	0.944055944
16	0.939285714
};
\addlegendentry{7F$_{\textrm{token}}$+1L};
\addplot[Style66c] table {
1	0.970771504
2	0.948059947
3	0.953070528
4	0.944795617
5	0.954959786
6	0.94207723
7	0.955334988
8	0.9527897
9	0.959654179
10	0.96484375
11	0.953917051
12	0.957792208
13	0.954918033
14	0.956790123
15	0.937062937
16	0.939285714
};
\addlegendentry{7F$_{\textrm{token}}$ (H8 disabled)};

\addplot[Style61a] table {
1	0.940772146
2	0.892219257
3	0.887905605
4	0.852928782
5	0.864343164
6	0.862849534
7	0.860215054
8	0.85944206
9	0.860230548
10	0.89453125
11	0.84562212
12	0.853896104
13	0.836065574
14	0.858024691
15	0.86013986
16	0.835714286
};
\addplot[Style61b] table {
1	0.938909231
2	0.896119893
3	0.884955752
4	0.857985672
5	0.876675603
6	0.873501997
7	0.865177833
8	0.856223176
9	0.873198847
10	0.875
11	0.84562212
12	0.831168831
13	0.860655738
14	0.851851852
15	0.846153846
16	0.825
};
\addplot[Style61c] table {
1	0.934348301
2	0.886881544
3	0.864306785
4	0.833965445
5	0.849329759
6	0.849533955
7	0.844499586
8	0.836909871
9	0.847262248
10	0.859375
11	0.81797235
12	0.824675325
13	0.840163934
14	0.808641975
15	0.797202797
16	0.778571429
};

\node[font=\small] at (axis cs: 7,0.97) {6+6 layers};
\node[font=\small] at (axis cs: 7,0.89) {6+1 layers};
\end{axis}
\end{tikzpicture}
\caption{Subject-verb agreement accuracies of the EN--DE models. The x-axis shows distances between the subject and the verb.}
\label{fig:longdist}
\end{figure}

The predefined attention patterns focus on relatively small local contexts. It could therefore be argued that the fixed attention models would perform worse on long-distance agreement, and that disabling the learned head in particular would be catastrophic.
We test this hypothesis by evaluating the models from Section~\ref{sec:results_mid} on the subject-verb agreement task of the English--German Lingeval test suite \cite{sennrich-2017-grammatical}. 
Figure~\ref{fig:longdist} plots the accuracies by distance between subject and verb. In the 6+6 layer configuration, no difference can be detected between the three examined scenarios. In the 6+1 layer configuration, the fixed-attention model does not seem to suffer from degraded results, whereas disabling the learned head leads to clearly lower results. This drop is due to the expected degradation of general translation quality (cf. Table~\ref{tab:ablation}, ablation study) and is not worse than the degradation observed by disabling one of the fixed local context heads.

%\textcolor{red}{this looks quite different from the comparable ACL graph...}

\subsection{Word sense disambiguation}

It has been shown that the encoder of Transformer-based MT models includes semantic information beneficial for WSD \cite{tang-etal-2018-self,tang-etal-2019-encoders}. In this respect, a model with predefined fixed patterns may struggle to encode global semantic features. To this end, we evaluate our models on two German--English WSD test suites, ContraWSD \cite{rios-gonzales-etal-2017-improving} and MuCoW \cite{raganato-etal-2019-mucow}.\footnote{As MuCoW is automatically built using various parallel corpora, we discarded those ones included in our training. We only report the average result from the TED \cite{cettolo2013report} and Tatoeba \cite{tiedemann-lrec12} sources.}
%and some of them are included also in our training, we only report the average result from the TED \cite{cettolo2013report} and Tatoeba \cite{tiedemann-lrec12} sources.}

Table~\ref{tab:wsd} shows the performance of our models on the WSD benchmarks.
Overall, the model with 6 decoder layers and fixed attentive patterns (7F$_\textrm{token}$+1L) achieves higher accuracy than the model with all learnable attention heads (8L), while the 1-layer decoder models show the opposite effect. It appears that having 6 decoder layers can effectively cope with WSD despite having only one learnable attention head. Interestingly enough, when we disable the learnable attention head (7F$_\textrm{token}$ H8 disabled), performance drops consistently in both test suites, showing that the learnable head plays a key role for WSD, specializing in semantic feature extraction.  

\begin{table}
\centering
\begin{adjustbox}{max width=\linewidth}
\setlength{\tabcolsep}{4pt}
\begin{tabular}{lrrrr}
\toprule
& \multicolumn{2}{c}{ContraWSD} & \multicolumn{2}{c}{MuCoW} \TBstrut\\
\cmidrule(lr){2-3} \cmidrule(lr){4-5}
Encoder heads & 6+1 & 6+6 & 6+1 & 6+6 \TBstrut\\
\midrule
%MUCOW ALL accuracy
%All learned & \bf 0.804 & 0.831 & \bf 0.615 & 0.645\\
%7 fixed, 1 learned & 0.793 & \bf 0.834 & \bf 0.615 & \bf 0.652 \\
%7 fixed (8 disabled) & 0.761 & 0.816 & 0.611 & 0.642\\
%MUCOW -> average TED and Tatoeba
8L & \bf 0.804 & 0.831 & \bf 0.741 & 0.761 \TBstrut\\
7F$_\textrm{token}$+1L & 0.793 & \bf 0.834 &  0.734 & \bf 0.772 \TBstrut\\
7F$_\textrm{token}$ (H8 disabled) & 0.761 & 0.816 & 0.721 & 0.757 \TBstrut\\
\bottomrule
\end{tabular}
\end{adjustbox}
\caption{Accuracy scores of the German--English models on the ContraWSD and MuCoW test suites.}
\label{tab:wsd}
\end{table}

\section{Conclusion}

In this work, we propose to simplify encoder self-attention of Transformer-based NMT models by replacing all but one attention heads with fixed positional attentive patterns that require neither training nor external knowledge. 
%We build upon existing work on analyzing encoder self-attention \cite{voita-etal-2019-analyzing}, and introduce seven \textit{non-learnable} attention heads.

We train NMT models on different data sizes and language directions with the proposed fixed patterns, showing that the encoder self-attention can be simplified drastically, reducing parameter footprint at training time without degradation in translation quality. In low-resource scenarios, translation quality is even improved. Our extensive analyses show that i) only adjacent and previous token attentive patterns contribute significantly to the translation performance, ii) the trainable encoder head can also be disabled without hampering translation quality if the number of decoder layers is sufficient, iii) encoder attention heads based on locality patterns are beneficial in low-resource scenarios, but may affect the semantic feature extraction necessary for addressing lexical ambiguity phenomena.% in MT.

Apart from the consistent %and solid [the reviewers will tell us if they're solid or not :D]
results given by our simple fixed encoder patterns, this work opens up potential further research for simpler and more efficient neural networks for MT, such as synthetic self-attention patterns \cite{tay2020synthesizer}.

\section*{Acknowledgments}
%The authors wish to acknowledge CSC – IT Center for Science, Finland, for computational resources.
\vspace{1ex}
\noindent
\begin{minipage}{0.1\linewidth}
    \vspace{-10pt}
    \raisebox{-0.2\height}{\includegraphics[trim =32mm 55mm 30mm 5mm, clip, scale=0.2]{erc.ai}} \\
    \raisebox{-0.25\height}{\includegraphics[trim =0mm 5mm 5mm 2mm,clip,scale=0.078]{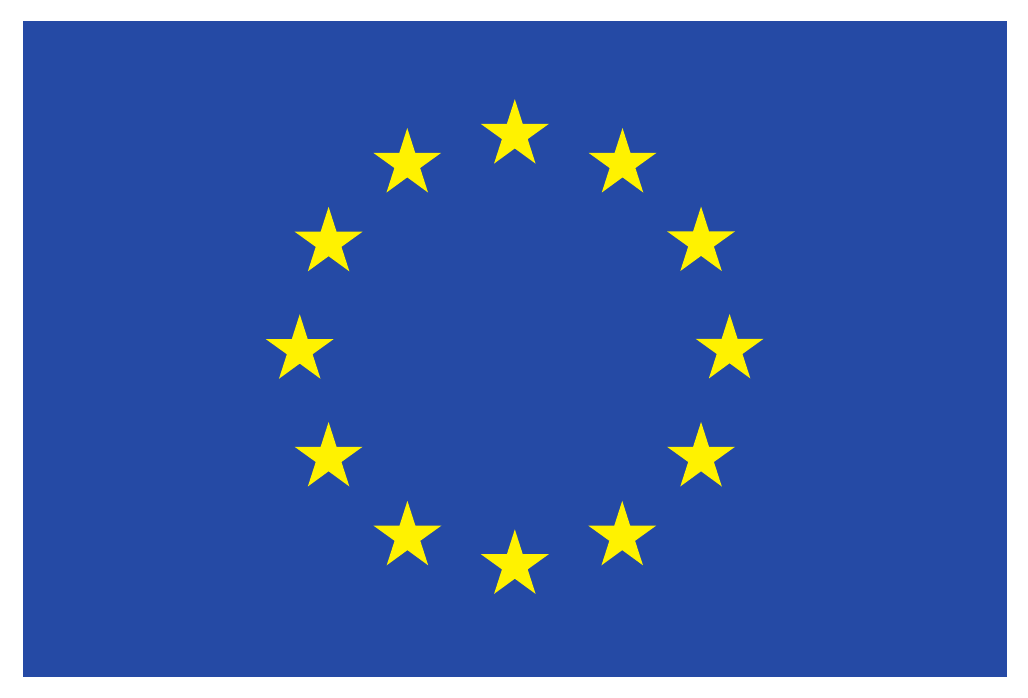}}
\end{minipage}
\hspace{0.01\linewidth}
\begin{minipage}{0.85\linewidth}
This work is part of the FoTran project, funded by the European Research Council (ERC) under the European Union's Horizon 2020 research and innovation programme (grant agreement No 771113).
  \vspace{1ex}
\end{minipage}

The authors gratefully acknowledge the support of the CSC – IT Center for Science, Finland, for computational resources.
%and project 270354/273457. 
%% project  270354/273457 is Anssi machine
Finally, We would also like to acknowledge NVIDIA and their GPU grant.

\bibliographystyle{acl_natbib}
\bibliography{emnlp2020}

\end{document}